\documentclass[3p,review,times]{elsarticle}

\usepackage[figuresright]{rotating}
\usepackage{amsopn}
\usepackage{amssymb}
\usepackage{amsmath}
\usepackage{subfig}
\usepackage{array}
\usepackage{tabularx}
\usepackage{multirow}
\usepackage{booktabs}
\usepackage{graphicx}
\usepackage{mathtools}
\usepackage{algorithmic}
\usepackage{algorithm}
\usepackage{listings}
\usepackage{setspace}
\usepackage{url}
\usepackage{float}
\usepackage{epsf} 
\usepackage{tikz}
\usepackage{amsthm}
\usepackage{color}
\usepackage{lineno}

\theoremstyle{definition}
\usepackage{url}

\newcommand{\etal}{\textit{et al.}}
\newcommand{\ie}{\textit{i.e.}}
\newcommand{\eg}{\textit{e.g.}}
\newcommand{\viz}{\textit{viz.}}
\newcommand{\mI}{\mathbf{I}}
\newcommand{\mone}{\mathbf{1}}
\newcommand{\fig}[1]{Figure \ref{#1}}
\newcommand{\eq}[1]{Eqn. (\ref{#1})}
\newcommand{\sect}[1]{Section \ref{#1}}
\newcommand{\tab}[1]{Table \ref{#1}}

\DeclareGraphicsExtensions{.pdf,.png}
\graphicspath{{../images/}}

\makeatletter
\newcommand{\pushright}[1]{\ifmeasuring@#1\else\omit\hfill$\displaystyle#1$\fi\ignorespaces}
\newcommand{\pushleft}[1]{\ifmeasuring@#1\else\omit$\displaystyle#1$\hfill\fi\ignorespaces}
\makeatother

\DeclareGraphicsExtensions{.pdf}
\graphicspath{{./}}

\begin{document}

\begin{frontmatter}




\title{Product Graph-based Higher Order Contextual Similarities for Inexact Subgraph Matching}

\author[cvc]{Anjan Dutta\corref{cor1}}
\ead{adutta@cvc.uab.es}
\author[cvc]{Josep Llad\'{o}s}
\ead{josep@cvc.uab.es}
\author[iam]{Horst Bunke}
\ead{bunke@iam.unibe.ch}
\author[isi]{Umapada Pal}
\ead{umapada@isical.ac.in}
\cortext[cor1]{Corresponding author}
\address[cvc]{Computer Vision Center, Universitat Aut\`{o}noma de Barcelona, Edifici O, Campus UAB, 08193 Bellaterra, Barcelona, Spain}
\address[iam]{Institute of Computer Science, University of Bern, Neubr\"{u}ckstrasse 10, CH-3012 Bern, Switzerland}
\address[isi]{Computer Vision and Pattern Recognition Unit, Indian Statistical Institute, 203, B.T.Road, Kolkata-108, India}

\begin{abstract}
Many algorithms formulate graph matching as an optimization of an objective function of pairwise quantification of nodes and edges of two graphs to be matched. Pairwise measurements usually consider local attributes but disregard contextual information involved in graph structures. We address this issue by proposing contextual similarities between pairs of nodes. This is done by considering the tensor product graph (TPG) of two graphs to be matched, where each node is an ordered pair of nodes of the operand graphs. Contextual similarities between a pair of nodes are computed by accumulating weighted walks (normalized pairwise similarities) terminating at the corresponding paired node in TPG. Once the contextual similarities are obtained, we formulate subgraph matching as a node and edge selection problem in TPG. We use contextual similarities to construct an objective function and optimize it with a linear programming approach. Since random walk formulation through TPG takes into account higher order information, it is not a surprise that we obtain more reliable similarities and better discrimination among the nodes and edges. Experimental results shown on synthetic as well as real benchmarks illustrate that higher order contextual similarities add discriminating power and allow one to find approximate solutions to the subgraph matching problem.
\end{abstract}


\end{frontmatter}

% \linenumbers

\section{Introduction}
\label{sec:intro}
The use of representational models allowing to describe visual objects in terms of their components and their relations has gained interest among the research community. Relations between parts express a notion of context, so objects can be recognized even if they appear in highly distorted or noisy visual conditions. At the heart of structural models, graphs have been used over decades as robust and theoretically solid representations. When objects are represented structurally by attributed graphs, their comparison for recognition and retrieval purposes is performed using some form of \emph{graph matching}. Basically, an \emph{attributed graph} is a 4-tuple $G(V,E,\alpha,\beta)$ comprising a set $V$ of \emph{vertices} or \emph{nodes} together with a set $E\subseteq V\times V$ of \emph{edges} and two \emph{mappings} $\alpha:V\rightarrow \mathbb{R}^m$ and $\beta:E\rightarrow \mathbb{R}^n$ which respectively assign attributes to the nodes and edges. Given two attributed graphs, \emph{graph matching} can roughly be defined as a process of finding a mapping between the node and edge sets of two graphs that satisfies some constraints. Graph matching is a very fundamental problem in theoretical computer science and it has been successfully applied to many applications in computer vision and pattern recognition~\cite{LladosPAMI2001,Carcassoni2003,Hays2006,Brendel2011,Duchenne2011,Jiang2011,Li2011,Dutta2013}. In fact, graphs are used in several domains including chemical compound, social network, biological and protein structure, program flow etc. Any complex data structure can efficiently be represented with a graph in terms of relations among different entities. Hence the theory of graph matching can be applied to different problems in any of the previously mentioned fields.

Graph matching is a very well known but challenging task. It requires high time complexity to be solved in an exact way. A particular class of algorithms resulting from the consideration of outliers\footnote{By outliers, we mean the set of nodes and edges in the target graph that are not related to nodes and edges of the pattern graph.} is called \emph{subgraph matching}. Roughly, it can be defined as matching one graph (pattern graph) as part of the other (target graph). Addressing this problem is justified since in computer vision and pattern recognition applications, due to the existence of background, occlusions and geometric transformations of objects, outliers do exist in many scenarios. Most of the modern algorithms formulate (sub)graph matching as an optimization of an objective function of pairwise quantification of the nodes and edges. But pairwise measurements usually consider local attributes and do not encode context information \ie~neighbourhood information of nodes and edges. As a result, pairwise quantifications are less reliable, especially in case of subgraph matching scenarios~\cite{Yang2013}. This is because of the existence of large amounts of outliers, which most of the graph matching algorithms do not take into account~\cite{Umeyama1988,Zaslavskiy2009,Yang2015}. This fact is also reported for many real world scenarios where higher order similarities that employ contextual information have been proposed as a solution~\cite{Yang2012}. We have also experienced quite similar phenomena with pairwise measurements in the presence of outliers. This fact will be presented in the experimental study in \sect{sec:expt} (symbol spotting experiment) as a benefit of the proposed contextual similarities (see \fig{fig:matching-bed}).

\begin{figure*}[!t]
\begin{center}
\includegraphics[width=\textwidth]{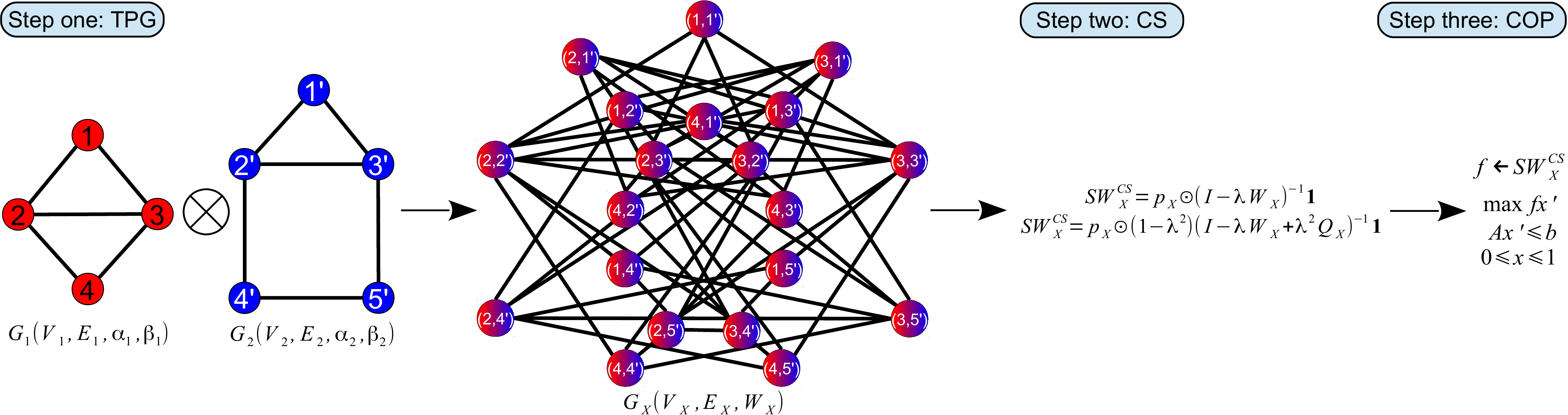}
\caption{Outline of the proposed method: Step one: computation of the tensor product graph (TPG) $G_X$ of two operand graphs $G_1$ and $G_2$, Step two: algebraic procedure to obtain contextual similarities (CS), Step three: constrained optimization problem (COP) for subgraph matching.}
\label{fig:outline}
\end{center}
\end{figure*}

The main motivation of this work is the use of contextual information of nodes (\ie~neighbouring structures) to make subgraph matching more robust and efficient in large scale visual recognition scenarios. A second key component is the formulation of subgraph matching with approximate algorithms. Therefore, in this paper we propose an inexact subgraph matching methodology based on \emph{tensor product graph} (TPG). Given two attributed graphs it is quite straight forward to get the pairwise similarities and assign them as weights on the edges of TPG (step one in \fig{fig:outline}). Next one can think of having a random walk from node to node considering the weights on the edges as the probabilities to proceed to the next node. Finally, we accumulate the probabilities of finishing a walk at each of the vertices, which we refer to as \emph{contextual similarities} (CS), where the context of each node is the set of its neighbouring nodes. This procedure essentially diffuse the pairwise similarities in the context of neighbouring nodes. This information can be obtained by simple algebraic operation of the adjacency (or weight) matrix of the product graph (step two in \fig{fig:outline}). A similar phenomenon is termed \emph{diffusion on tensor product graph}, which is well known to capture the geometry of data manifold and higher order contextual information between objects~\cite{Coifman2006,Yang2012}. We formulate subgraph matching as a node and edge selection problem in TPG. To do that we use those contextual similarities and formulate a constrained optimization problem (COP) to optimize a function constructed from the higher order similarity values (step three in \fig{fig:outline}). We solve the optimization problem with linear programming (LP) which is solvable in polynomial time. Depending on whether we need a discrete solution or not, we may add a discretization step. In this paper we will show that the higher order contextual similarities allow us to relax the constrained optimization problem in real world scenarios.

The main contribution of this work is to propose a random walk based approach to obtain contextual similarities between nodes that capture higher order information. In the literature, there are methods that formulate graph matching explicitly using higher order information of nodes and edges\cite{Lee2011,Duchenne2011a}. They often embed the higher order relations of nodes by hypergraph matching formulations. Usually the affinity of feature tuples are modelled using tensor and later approximate algorithms are proposed to obtain the correspondences~\cite{Lee2011}. Our work takes into account the higher order context information by propagating the pairwise similarities with random walk formulation. The procedure counts the number of incoming walks to the nodes of TPG. Here the general hypothesis is that a particular node in TPG, that constitutes a pair of nodes of the operand graphs likely to be matched, should receive a reasonably high amount of random walks. This idea is also supported by other works \cite{Szummer2002,Coifman2006,Cho2010,Lee2011,Yang2012,Cho2012}, where the idea of random walk has been used for obtaining similarity of pairs of nodes. Yet, our formulation and interpretation are completely different than the others. We have used the classical algebraic formulation to obtain the similarities between nodes whereas most of the previously mentioned algorithms have used an approximated iterative procedure to obtain the similarities or the node to node correspondences depending on the requirement. Apart from that, we have revealed that backtrackless walks~\cite{Aziz2013} can also be utilised for obtaining contextual similarities of pairs of nodes. Later, we use these contextual similarities to formulate a subgraph matching procedure as a node and edge selection procedure in TPG. This is an optimization problem and here we are motivated by the ILP formulation of Le Bodic~\etal~\cite{LeBodic2012}. We have proved the effectiveness of the proposal with a detailed experimental study.

The rest of the paper is organized in four sections. In \sect{sec:lit}, we give a formal definition of graph matching and present the state-of-the-art of (sub)graph matching algorithms. We present a product graph based subgraph matching methodology which includes the description of the contextual similarities in \sect{sec:context-simils} and the formulation of subgraph matching as an constrained optimization problem in \sect{sec:subgraph-matching}. In \sect{sec:expt} we present experimental results showing the robustness of the method. At last in \sect{sec:concl} we conclude the paper and outline possible future directions of this work.
%------------------------------------------------------------------------------------------------------------------------------------------------------------------------------------------------------------
\section{Previous work on graph matching}
\label{sec:lit}
\subsection{Formal definition of graph matching}
\label{ssec:def-gm}
Let $G_1(V_1,E_1,\alpha_1,\beta_1)$ and $G_2(V_2,E_2,\alpha_2,\beta_2)$ be two attributed graphs with $|V_1|=n_1$ and $|V_2|=n_2$. Let $\kappa^{v}_{i_1i_2}$ measures the affinity between the nodes $u_{i_1}\in V_1$, $u_{i_2}\in V_2$ and $\kappa^{e}_{i_1i_2,j_1j_2}$ measures the affinity between the edges $(u_{i_1},v_{j_1})\in E_1$ and $(u_{i_2},v_{j_2})\in E_2$. Given these two graphs together with node and edge affinity measures, one can compute the affinity matrix $\mathbf{K}\in \mathbb{R}^{n_1n_2\times n_1n_2}$ in a concise way as follows:
\begin{equation}
\mathbf{K}(i_1i_2,j_1j_2)=\begin{cases}
    \kappa^{v}_{i_1i_2}, & \text{if }i_1=j_1\text{ and }i_2=j_2.\\
   	\kappa^{e}_{i_1i_2,j_1j_2}, & \text{if }(u_{i_1},v_{j_1})\in E_1\text{ and}\\&(u_{i_2},v_{j_2})\in E_2.\\
    0, & \text{otherwise}.
  \end{cases}
  \label{eqn:gm}
\end{equation}
Using $\mathbf{K}$, graph matching can be concisely formulated as the following quadratic assignment problem (QAP):
\begin{equation}
\max_{\mathbf{X}\in\Pi}\hspace{5mm}J_{gm}(\mathbf{X}) = \text{vec}(\mathbf{X})^T\mathbf{K}\text{vec}(\mathbf{X})
\label{eqn:qap}
\end{equation}
where $\mathbf{X}\in \Pi$ is constrained to be a one-to-one mapping, \ie~$\Pi$ is the set of partial permutation matrices:

\begin{equation}
\Pi = \lbrace\mathbf{X}\in \lbrace0,1\rbrace^{n_1\times n_2}:\mathbf{X}\mone_{n_2}=\mone_{n_1}, \mathbf{X}^T\mone_{n_1}=\mone_{n_2}\rbrace
\label{eqn:perm-cons}
\end{equation}
Solving \eq{eqn:qap} in an exact way leads to IQP which is a typical NP-hard problem.

\subsection{Related works}
\label{ssec:rel-works}
Very extensive and detailed reviews of all kind of graph matching methods for pattern recognition is available in~\cite{Conte2004,Foggia2014,Vento2015}. In pattern recognition, the research community has divided the graph matching methods into two broad categories: (1) \emph{exact} and (2) \emph{inexact} graph matching. Exact matching requires a strict correspondence between the two objects being matched or at least between their sub parts~\cite{Ullman1976,Cordella1999,Cordella2004}. However, the strong constraints of the exact matching algorithms are too rigid for real applications in computer vision. This is because the description of an object (or part of it) represented as the node/edge attributes of a graph is a vector resulting from some statistical analysis. For that reason researchers proposed inexact graph matching which allows some tolerance under a similarity model and in this case a matching can occur when two graphs being matched are structurally different to some extent. Among the earlier works, a group of methods worked with \emph{tree search with backtracking}~\cite{Tsai1979,Tsai1983}. These methods gave birth of the popular \emph{graph edit distance}. In these papers the authors introduced the \emph{graph edit costs} \ie~the costs for substituting, deleting, inserting and merging nodes and edges to define a distance between two graphs~\cite{Bunke1997,Bunke1999}. Later, graph edit distance was shown to be effective for \emph{explicit graph embedding}~\cite{Bunke2012} and \emph{graph kernel}~\cite{Neuhaus2007}, although detailed discussions on these topics are out of the scope of the article.

Most of the modern methods address the inexact matching problem and focus on the relaxation of permutation constraints (\eq{eqn:perm-cons}). Based on the approximation strategy and the applied methodology, these methods can be broadly classified into three categories: (1) \emph{spectral decomposition}, (2) \emph{semidefinite programming (SDP) relaxation} and (3) \emph{convex relaxation}.

\subsubsection{Spectral decomposition}
The first set of methods estimates the graph permutation matrix $\mathbf{X}$ with an orthogonal one, \ie, $\mathbf{X}^T\mathbf{X}=\mI$. Under this orthogonal constraint, optimizing graph matching can be solved as an eigenvalue problem~\cite{Umeyama1988,Caelli2004}. These methods often suffer when the spectral embeddings of the adjacency matrices are not uniquely defined. The problems in computer vision are much more complex and to deal with them several researchers have proposed a wealth of effective and efficient solutions~\cite{Leordeanu2005,Cour2006}.

\subsubsection{Semidefinite programming}
Semidefinite programming is a general relaxation tool for approximating many combinatorial problems. Consequently, it has largely been used to approximate the graph matching problem. These methods usually redefine the objective of graph matching by introducing a new variable $\mathbf{Y}\in\mathbb{R}^{n_1n_2\times n_1n_2}$ such that $\mathbf{Y}=\text{vec}(\mathbf{X})\text{vec}(\mathbf{X})^T$. Here semidefinite programming (SDP) approximates the non-convex constraint on $\mathbf{Y}$ as a semidefinite one so that it satisfies $\mathbf{Y}-\text{vec}(\mathbf{X})\text{vec}(\mathbf{X})^T\ge 0$. The main advantage of the SDP methods is their theoretical assurance to obtain a solution in polynomial time. After computing $\mathbf{Y}$ using SDP, a probabilistic post-processing step is included to get a discrete solution~\cite{Torr2003,Schellewald2005}. Nonetheless, sometime SDP algorithms are rather expensive as the size of $\mathbf{Y}$ is quadratic to the actual problem size.

\subsubsection{Convex relaxation}
Most of the other methods relax $\mathbf{X}\in\mathcal{D}$ to be a doubly stochastic matrix, the convex hull of the set of permutation matrices:

\begin{equation}
\mathcal{D}=\lbrace\mathbf{X}\in \mathbb{R}^{n_1\times n_2}:\mathbf{X}\ge 0, \mathbf{X}\mone_{n_2}=\mone_{n_1}, \mathbf{X}^T\mone_{n_1}=\mone_{n_2}\rbrace
\end{equation}

Presuming this constraint, optimizing graph matching can be considered as a convex quadratic programming problem. Various methods have already been proposed to find a local optimum. One of the first approaches along this line was proposed by Almohamad and Duffuaa~\cite{Almohamad1993}, where the authors transformed the quadratic optimization problem to a linear one and solved it with a combination of linear programming and the Hungarian algorithm. Gold and Rangarajan~\cite{Gold1996} proposed a graduated assignment algorithm by combining graduated nonconvexity, two-way assignment constraints, and sparsity. The method iteratively solves a series of linear approximations of the cost function using Taylor's series expansion to find a fast and accurate solution. Pelillo~\etal~\cite{Pelillo1999} proposed a series of algorithms that mainly search for a maximal clique in an association graph of two operand graphs for their matching. They utilised the replicator dynamics equation to solve the problem of maximal clique finding in an approximate way. Van Wyk and Van Wyk~\cite{VanWyk2004} proposed an algorithm to iteratively project the approximate correspondence matrix onto the convex domain. Similar to~\cite{Gold1996}, they enforced two-way assignment constraints but did not use elaborated penalty terms, graduated nonconvexity, or sophisticated annealing mechanisms to avoid poor local minima. Torresani~\etal~\cite{Torresani2008,Torresani2013} designed a complex energy function depending on the appearance, geometry and spatial coherence of the features to solve the matching task. They proposed a dual decomposition based approximate algorithm to optimize the energy function. Zass and Shashua~\cite{Zass2008} presented a convex relative entropy error model emerged from a probabilistic interpretation of the graph matching problem. Leordeanu~\etal~\cite{Leordeanu2009} proposed an integer projection algorithm with climbing and convergence properties to optimize the objective function in an integer domain. They claimed their algorithm gives excellent results starting from the solution returned by any graph matching algorithm. Zaslavskiy~\etal~\cite{Zaslavskiy2009} proposed a path following algorithm of a convex-concave problem obtained by linear interpolation of convex and concave formulations, starting from the convex relaxation. Inspired by the PageRank algorithm~\cite{Page1999}, Cho~\etal~\cite{Cho2010} proposed a random walk based iterative procedure for the graph matching problem. As an extension to~\cite{Zaslavskiy2009}, Liu~\etal~\cite{Liu2012a} proposed another path following strategies for the non-convex quadratic programming problem. Their algorithm takes into account the directed graph whereas the algorithm proposed by Zaslavskiy~\etal~\cite{Zaslavskiy2009} considered undirected graphs. Their algorithm is also applicable for getting the discrete solution from the continuous solution of any other graph matching algorithm. Zhou and De la Torre~\cite{Zhou2012,Zhou2013} presented a novel framework based on the factorization of the pairwise affinity matrix which provides several benefits from the graph matching point of view.

Apart from formulating better approximations for \eq{eqn:qap}, there are two more directions of graph matching methods that reported improved results: (1) learning the pairwise affinity matrix~\cite{Caetano2009,Leordeanu2012} and (2) explicitly incorporating higher order features~\cite{Zass2008,Lee2011,Duchenne2011a}. The first set of methods learns in a supervised or unsupervised way the compatibility functions such that the solution of the resulting graph matching problem best matches the expected solution that a human would manually provide. A slightly different approach is to learn class specific graph models to match with target graphs~\cite{Cho2013}. The second category of methods addresses the problem of establishing correspondences between two sets of visual features using higher order constraints instead of the unary or pairwise ones~\cite{Lee2011,Duchenne2011a}. By imposing higher order constraints, the authors formulate graph matching as a hypergraph matching task which is formulated as the optimization of a multilinear objective function defined by a tensor representing the affinity between feature tuples. There are methods that work with equal sized graphs~\cite{Zaslavskiy2009,Liu2012}. But most of the modern methods can deal with graphs of different size, though usual graph matching algorithms perform worse in subgraph matching scenario~\cite{Yang2012}.

Despite having many methods for graph matching available, very few of them explore the fact of context for strengthening node-node matching affinities~\cite{Cho2010}. Also many of the above methods fail to perform well in presence of a high number of outliers. To solve these problems, in this paper, we propose contextual similarities on the tensor product graph and use them for subgraph matching. 
%------------------------------------------------------------------------------------------------------------------------------------------------------------------------------------------------------------
\section{Contextual similarities on product graph}
\label{sec:context-simils}
A \emph{tensor product graph} (TPG) $G_X(V_X,E_X,p_X,W_X)$ of two operand graphs $G_1(V_1,E_1,\alpha_1,\beta_1)$ and $G_2(V_2,E_2,\alpha_2,\beta_2)$ is a graph such that the vertex set $V_X$ is the Cartesian product $V_1\times V_2$ \ie
\begin{align*}
V_X =\lbrace (u_i,u_j):u_i\in V_1, u_j\in V_2\rbrace
\end{align*}
and the edge set $E_X$ is defined as:
\begin{align*}
E_X =\lbrace ((u_i,u_j),(v_i,v_j)):(u_i,v_i)\in E_1\text{ and }(u_j,v_j)\in E_2, u_i,v_i\in V_1,u_j,v_j\in V_2\rbrace
\end{align*}
and $p_X$ and $W_X$ contain weights as a function of $\alpha_1$, $\alpha_2$ and $\beta_1$, $\beta_2$, respectively.

A very interesting property of the adjacency matrix $A$ of any graph $G$ is that the $(i,j)$th entry of $A^n$ denotes the number of walks of length $n$ from node $v_i$ to node $v_j$. One can bring the same analogy to an edge weighted graph where each edge is associated with weights in $[0,1]$. Let $W$ be such a matrix that contains weights on the edges. These weights can be considered as the probabilities (after normalization, when each of its columns sums to one) of a random walker moving from one node to another. Then following the same idea the $(i,j)$th entry of $W^n$ denotes the probability of having a walk of length $n$ from the node $v_i$ to the node $v_j$. Now, let $p_0$ be the stopping probability distribution over vertices ($p_0$ sums to one), then the probability distribution $p_k$ describing the location of the random walker at the end of $k$th transition is $p_{k}=p_{0}\odot W^{k}\mone$~\cite{Vishwanathan2010}. Here a higher entry reveals that more and more random walks finish at a particular node and this particular node is better connected with rest of the graph.

The same procedure can be simulated on a TPG of two operand graphs and can be used to capture higher order contextual information. The process can be started by assigning the pairwise similarities of the edges as the weights on the corresponding edges of TPG. Let $W_X$ be such a matrix, where each of its columns sums to one (with weights converted to probabilities). This matrix basically contains the probabilities of transition of a random walker from a node to the next one. Let $p_X$ be the stopping probability distribution over the vertices. Practically this is a vector which contains the node to node similarities that sum up to one. Then simultaneous walking can be performed from node to node taking these probabilities to move from one node to the next one. Similar to the above explanation, the position of the random walker, after finishing a $k$-length walk, can be obtained by a probability distribution $p_{X_k}=p_X\odot W_{X}^{k}\mone$. The $i$th component of $p_{X_k}$ denotes the probability of finishing a $k$-length walk at the vertex enumerated $i$.

Let
\begin{equation}
S_{p_X}=p_X\odot(\lim_{n\rightarrow \infty}\sum_{k=0}^n W_X^k) \mone
\end{equation}
Similar to the previous explanation, here the $i$th component of $S_{p_X}$ indicates the likelihood of finishing a walk at node $v_i$ and a higher entry reveals that the node is better connected with rest of the TPG. As the weights on the edges of TPG come from the similarities of edges of the operand graphs, here a better connected node of TPG is supposed to constitute a promising pair of matched nodes. Since the walking is performed through the edges of TPG, the accumulated weights take into account contextual information. This procedure of accumulating weights by considering connections between objects is proved to be more discriminative for objects where contextual information is important. This is also the intrinsic idea of \emph{graph diffusion}, which is well known to capture higher order context similarities and complex relations. It is observed by several recent approaches that when pairwise similarities are diffused in the context of other neighbouring points, more reliable similarities are obtained~\cite{Szummer2002,Coifman2006,Cho2010,Lee2011,Yang2012,Cho2012}. In other words, this random walk procedure reevaluates the pairwise similarities within the context of neighbouring nodes, resulting in more robust affinities that we termed \emph{contextual similarities}. In our graph matching module, we use these reevaluated similarity values, which are more robust than pairwise similarities. In this way our method is more advantageous than other existing graph matching methods based on random walk~\cite{Cho2010}. In this paper, we model the procedure of walking in two different ways, \viz, \emph{random walks} and \emph{backtrackless walks} which we describe below.

\subsection{Random walks}
\label{ssec:rw}
The easiest way to get the contextual similarities between pairs of nodes through a graph is by propagating the pairwise similarity information with random walks on TPG. Below we use $\mI$ to denote the identity matrix. When it is clear from context we will not mention the dimensions of any of the vectors or matrices. The process of obtaining contextual similarities with random walks can be defined as:

\begin{equation}
SW^{CS}_X = p_X\odot(\lim_{n\rightarrow \infty}\sum_{k=0}^n \lambda^k W_{X_k}) \mone= p_X\odot(\mI-\lambda W_X)^{-1}\mone
\label{eqn:rw1}
\end{equation}
where
\begin{equation}
W_{X_k} = W_X^k
\label{eqn:rw2}
\end{equation}
Here $\lambda$ is a weighting factor to discount the longer walks, as they often contain redundant or repeated information such as cycles. In this paper we always choose $\lambda=\frac{1}{a}$, where $a=\min(\Delta^+(W_X),\Delta^-(W_X))$. Here $\Delta^+(W_X)$, $\Delta^-(W_X)$ are respectively the maximum outward and inward degree of $W_X$~\cite{Gartner2003a}. The infinite summation of \eq{eqn:rw1} converges to $(\mI-\lambda W_X)^{-1}$ for sufficiently small values of $\lambda$.

\eq{eqn:rw1} can be efficiently computed by solving $(\mI-\lambda W_X)x=\mone$ by conjugate gradient methods which allows to avoid the expensive matrix inversion and then performing Hadamard product by $p_X$. An entry $W^{CS}_X(\omega_i,\omega_j)$ indicates the likelihood of finishing a random walk at the node $(\omega_i,\omega_j)$ of TPG. Since the weights on the edges of TPG are derived from the pairwise similarities of edge attributes, a higher value in $W^{CS}_X(\omega_i,\omega_j)$ reveals a better candidate for pairwise matching. Here it is to be noted that $W^{CS}_X$ is a column vector, the comma separated double subscript is used just to ease the understanding.

For the sake of understandability, let us illustrate the idea using a simple example. Let us consider an edge similarity matrix $W$ (\eq{eqn:ex-mat-walks}) and node similarity vector $p$ (\eq{eqn:ex-vect-walks}) as follows:

\noindent
\begin{minipage}{.5\textwidth}
\begin{equation}
W = 
\begin{bmatrix}
0			&w^{21}		&w^{31}\\
w^{12}		&0			&w^{32}\\
w^{13}		&w^{23}		&0
\end{bmatrix}
\label{eqn:ex-mat-walks}
\end{equation}
\end{minipage}
\begin{minipage}{.5\textwidth}
\begin{equation}
p = 
\begin{bmatrix}
p^1\\		
p^2\\
p^3
\end{bmatrix}
\label{eqn:ex-vect-walks}
\end{equation}
\end{minipage}
Here the contextual similarities obtained after finishing the 2-length walks are
\begin{equation}
%\resizebox{0.9\textwidth}{!}{$%
p\odot(\mI+W_1+W_2)\mone=
\begin{bmatrix}
p^1(1+w^{12}w^{21}+w^{13}w^{31})+p^1(w^{21}+w^{23}w^{31})+p^1(w^{31}+w^{32}w^{21})\\
p^2(w^{12}+w^{13}w^{32})+p^2(1+w^{21}w^{12}+w^{23}w^{32})+p^2(w^{32}+w^{31}w^{12})\\
p^3(w^{13}+w^{12}w^{23})+p^3(w^{23}+w^{21}w^{13})+p^3(1+w^{31}w^{13}+w^{32}w^{23})\\
\end{bmatrix}%$%
%}%
\label{eqn:ex-mult-walks}
\end{equation}
The rows of the matrix on RHS of \eq{eqn:ex-mult-walks} show the contextual similarities of respective nodes. As the entries of $W$ contain the edge similarities, it is clear that the procedure takes into account the similarities of the walks of different length (walks upto length two in this example) to respective nodes. Hence it is evident that the exponentiation plus summation procedure takes into account information from the context to determine the strength of a particular node. An actual occurrence of a pattern graph in the target graph creates higher similarities in the neighbourhood, which clearly effects the connected nodes. This formulation enhances the pairwise similarities with more information from context. 

The main drawback of the random walk based procedure is that it backtracks an edge in case of a undirected graph and this reduces its discriminatory power. To solve this limitation, \emph{backtrackless walks}, a variation of random walks have been recently proposed~\cite{Aziz2013}. In the next section we describe how to adapt this idea to our approach.

\subsection{Backtrackless walks}
\label{sssec:btlw}
A similar formulation as random walks can be used with \emph{backtrackless walks}~\cite{Aziz2013}. The backtrackless walks are random walks but do not backtrack an edge and for that a variation of exponentiation of the weight matrix is available~\cite{Stark1996}. The main advantage of backtrackless walks over random walks is that they avoid tottering, and hence increase the discriminative power of the resulting contextual similarities. The benefit of contextual similarities generated through this type of walks has been corroborated in our experiments on undirected graphs (see~\sect{ssec:expt-cmu} and \sect{ssec:expt-ss}). The process of obtaining contextual similarities with backtrackless walks can be defined as (proof is provided in the supplemental material):

\begin{align}
SW^{CS}_X&=p_X\odot(\lim_{n\rightarrow \infty}\sum_{k=0}^n \lambda^k W_{X_k})\mone \label{eqn:btlw1:1}\\
		 &=p_X\odot(1-\lambda^2)(\mI - \lambda W_X+\lambda^{2} Q_X)^{-1}\mone \label{eqn:btlw1:2}
\end{align}
where
\begin{equation}
W_{X_k} =
\begin{cases}
	W_X					& \text{if } k = 1 \\
	W_X^2-(Q_X+I)			& \text{if } k = 2 \\
   	W_{X_{k-1}}W_X-W_{X_{k-2}}Q_X   	& \text{if } k\geq 3
\end{cases}
\label{eqn:btlw2}
\end{equation}
Here $Q_X$ is a diagonal matrix where the $i$th or $(i,i)$th element is equal to the $i$th or $(i,i)$th element of $W_X^2$ minus one. Parameter $\lambda$ serves the same purpose as before. The summation in \eq{eqn:btlw1:1} converges for sufficiently small value of $\lambda$. Similar to \eq{eqn:rw1}, \eq{eqn:btlw1:2} can also be computed by solving $(\mI - \lambda W_X+\lambda^{2} Q_X)x=p_X$ and then multiplying the solution by $(1-\lambda^2)$.

Here the phenomena regarding context can be explained with the same example as in \eq{eqn:ex-mat-walks}, \eq{eqn:ex-vect-walks}. According to that example, $Q_X$ can be written as follows:
\begin{equation}
%\resizebox{0.9\textwidth}{!}{$%
Q_X =
\begin{bmatrix}
w^{12}w^{21}+w^{13}w^{31}-1	&0								&0\\								
0							&w^{12}w^{21}+w^{23}w^{32}-1		&0\\		
0							&0							&w^{13}w^{31}+w^{23}w^{32}-1\\
\end{bmatrix}%$%
%}%
\end{equation}

Now contextual similarities obtained after finishing the 2-length walks are:
\begin{equation}
%\resizebox{0.9\textwidth}{!}{$%
p_X\odot(\mI+W_1+W_2)\mone=
\begin{bmatrix}
p^1+p^1(w^{21}+w^{23}w^{31})+p^1(w^{31}+w^{32}w^{21})\\
p^2(w^{12}+w^{13}w^{32})+p^2+p^2(w^{32}+w^{31}w^{12})\\
p^3(w^{13}+w^{12}w^{23})+p^3(w^{23}+w^{21}w^{13})+p^3
\end{bmatrix}%$%
%}%
\end{equation}

Here also it is clear that the exponentiation plus the summation procedure takes into account contextual information to determine the strength of a particular node. In each iteration it eliminates the tottering effect (by eliminating the loops) with the special algebraic formulation in \eq{eqn:btlw2}. An actual occurrence of a pattern graph in the target graph creates higher pairwise similarities in the neighbourhood, and this also effects the connected nodes. This formulation enhances the pairwise similarities with more information from context/connection. To the best of our knowledge, this work is the first to explore the contextual similarities obtained by backtrackless walks. The contextual similarities obtained by these procedures are based on counting the common random or backtrackless walks in the operand graphs. As the edges of the product graph encode the similarity based on the features that are invariant to transformation, rotation and scale, contextual similarities obtained by these procedures are also invariant to such geometric transformation.

We use the contextual similarities obtained in the steps explained above to formulate a subgraph matching algorithm as a constrained optimization problem.

\section{Subgraph matching as a constrained optimization problem}
\label{sec:subgraph-matching}
The contextual similarities obtained in the previous step just provide node-node, edge-edge affinities and do not yield correspondences. We formulate subgraph matching as a node and edge selection problem in tensor product graph. To do that we use the contextual similarities obtained in the previous step to construct a maximization problem and solve it with a \emph{linear programming} formulation.
Let
\[
S_V(u_i,u_j) = W^{CS}_X(u_i,u_j),\text{  }(u_i,u_j)\in V_X
\]
and
\begin{align*}
S_E((u_i,u_j),(v_i,v_j)) = \frac{W^{CS}_X(u_i,u_j)}{\Delta^+(u_i,u_j)}+\frac{W^{CS}_X(v_i,v_j)}{\Delta^+(v_i,v_j)},(u_i,u_j), (v_i,v_j)\in V_X
\end{align*}
Here $\Delta^+(u_i,u_j)$ denotes the maximum outward degree of the node $(u_i,u_j)\in V_X$. Here it is to be clarified that both $S_V$ and $S_E$ are vectors, the comma separated double subscripts are just to ease understanding. Now clearly the dimension $S_V$ is $|V_X|$ and that of $S_E$ is $|E_X|$. We formulate the subgraph matching problem as a node and edge selection problem in the product graph. This can be regarded as a constrained optimization problem which maximize a function of higher order similarities of the nodes and edges. We construct the objective function as follows:

\begin{equation}
f(x,y) = S_V'x+S_E'y
\label{eqn:optm}
\end{equation}
where $x$ and $y$ are vectors containing variables denoting the probabilities of matching (selecting) the nodes and edges in the product graph respectively. For example, $x_{u_i,u_j}$ denote the probability of matching node $u_i\in V_1$ with node $u_j\in V_2$. Similarly, $y_{u_iu_j,v_iv_j}$ denote the probability of matching the edge $(u_i,v_j)\in E_1$ with edge $(u_j,v_j)\in E_2$. Both $x$ and $y$ contain probabilities, as the optimization problem in \eq{eqn:optm} is solved with linear or continuous programming with the domain $[0,1]$. To get a discrete values of the solution we apply the Hungarian algorithm on the node to node matching probabilities (here $x$). It is to be mention that the above objective function in \eq{eqn:optm}~can easily be written in the form of \eq{eqn:qap}, where $\mathbf{K}$ is given as follows:
\begin{equation*}
\mathbf{K}=
\begin{bmatrix}
S_V &0\\
0 &S_E
\end{bmatrix}
\end{equation*}

%TODO Mention that S_V is written as a diagonal matrix
Now let us introduce a set of constraints on the variables to satisfy the maximal common subgraph matching problem between the operand graphs $G_1$ and $G_2$ in TPG $G_X$.

\begin{itemize}
\item \textbf{Pattern node constraint} Each node $u_i\in V_1$ can be matched at most with one node $u_j\in V_2$ \ie~there can be at most a single node $(u_i,u_j)\in V_X$ for each $u_i\in V_1$.
\[
\sum_{u_2\in V_2}x_{u_i,u_j} <= 1,\forall u_i\in V_1
\]
\item \textbf{Pattern edge constraint} Each edge $(u_i,v_i)\in E_1$ can be matched at most with one edge $(u_j,v_j)\in E_2$ \ie~there can be at most a single edge $((u_i,u_j),(v_i,v_j))\in E_X$ for each $(u_i,v_i)\in E_1$
\[
\sum_{(u_j,v_j)\in E_2}y_{u_iu_j,v_iv_j} <= 1,\forall (u_i,v_i)\in E_1
\]
\item \textbf{Target node constraint} Each node $u_j\in V_2$ can be matched with at most one node in $u_i\in V_1$ \ie~there can be at most one node $(u_i,u_j)\in V_X$ for each $u_j\in V_2$.
\[
\sum_{u_i\in V_1}x_{u_i,u_j} <= 1,\forall u_j\in V_2
\]
\item \textbf{Target edge constraint} Each edge $(u_j,v_j)\in E_2$ can be matched with at most one edge $(u_i,v_i)\in E_1$ \ie~there can be at most one edge $((u_i,u_j),(v_i,v_j))\in E_X$ for each $(u_j,v_j)\in E_2$
\[
\sum_{(u_i,v_i)\in E_1}y_{u_iu_j,v_iv_j} <= 1,\forall (u_j,v_j)\in E_2
\]
\item \textbf{Outward degree constraint} The outward degree of a node $(u_i,u_j)\in V_X$ is bounded above by the minimum of the outward degree of the node $u_j\in V_1$ and $u_j\in V_2$ \ie~the number of outgoing edges from the node $(u_i,u_j)\in V_X$ is less than or equal to the minimum of the outward degree of the nodes $u_i\in V_1$ and $u_j\in V_2$.
\begin{align*}
\sum_{(v_i,v_j)\in V_X}y_{u_iu_j,v_iv_j} <= x_{u_i,u_j}.\min{(\Delta^+(u_i),\Delta^+(u_j))},\forall(u_i,u_j)\in V_X
\end{align*}
\item \textbf{Inward degree constraint} The inward degree of a node $(v_i,v_j)\in V_X$ is bounded above by the minimum of the inward degree of the node $v_i\in V_1$ and $v_j\in V_2$ \ie~the number of incoming edges to the node $(v_i,v_j)\in V_X$ is less than or equal to the minimum of the inward degree of the nodes $v_i\in V_1$ and $v_j\in V_2$.
\begin{align*}
\sum_{(u_i,u_j)\in V_X}y_{u_iu_j,v_iv_j} <= x_{v_i,v_i}.\min{(\Delta^-(v_i),\Delta^-(v_j))},\forall(v_i,v_j)\in V_X
\end{align*}
\item \textbf{Domain constraint} Finally, we restrict all the variables to be in between $[0,1]$.
\[
x_{u_i,u_j}\in [0,1],\text{  }\forall (u_i,u_j)\in V_X
\]
\[
y_{u_iu_j,v_iv_j}\in [0,1],\text{  }\forall ((u_i,u_j),(v_i,v_j))\in E_X
\]
\end{itemize}

The procedure of obtaining contextual similarities using random walks has some resemblance with the PageRank algorithm~\cite{Page1999}, power iteration method for eigenvalue problem~\cite{Duchenne2011a} or some other similar algorithms~\cite{Cho2010}. But their further utilization and the idea of backtrackless walks instead of simple random walks were not done before.
%------------------------------------------------------------------------------------------------------------------------------------------------------------------------------------------------------------
%\section{Complexity}
%\label{sec:comp}
%------------------------------------------------------------------------------------------------------------------------------------------------------------------------------------------------------------
\section{Experimental results}
\label{sec:expt}
In this section, we report four different experiments on different benchmark datasets and provide a comparison of the proposed algorithms with several state-of-the-art methods. In the first three experiments, we examine the performance of graph matching algorithms. We compare with eleven state-of-the-art algorithms: (1) Graduated assignment (GA)~\cite{Gold1996}, (2) Spectral matching (SM)~\cite{Leordeanu2005}, (3) Spectral matching with affine constraints (SMAC)~\cite{Cour2006}, (4) Integer projected fixed point method (IPFP-U)~\cite{Leordeanu2009}, (5) IPFP-S~\cite{Leordeanu2009}, (6) Probabilistic matching (PM)~\cite{Zass2008}, (7) Reweighted random walks matching (RRWM)~\cite{Cho2010}, (8) Factorized graph matching (FGM-D)~\cite{Zhou2012}, (9) FGM-U~\cite{Zhou2012}, (10) Dual decomposition (DD)~\cite{Torresani2013}, (11) High-order graph matching (HOGM)~\cite{Duchenne2011a}. Here IPFP-U and IPFP-S are basically the same algorithms; they only differ in the initialization procedure. Similary, FGM-U and FGM-D are also the same but designed for undirected and directed graphs, respectively. We have used some acronyms for referring to our algorithms as follows: (1) the algorithm working with pairwise similarities as PG-N, (2) the algorithm working with contextual similarities based on random walks as PG-R and (3) the algorithm working with contextual similarities based backtrackless walks as PG-B. For these experiments, we computed the matching \emph{accuracy} (\eq{eqn:acc}) and normalized \emph{objective score} (\eq{eqn:obj}) for evaluating and comparing the performance. The matching \emph{accuracy} is defined as the number of consistent matches between the corresponding matrix $\mathbf{X}_{algthm}$ given by an algorithm and ground truth $\mathbf{X}_{truth}$:
\begin{equation}
acc=\frac{tr(\mathbf{X}^{T}_{algthm}\mathbf{X}_{truth})}{tr(\mone_{|V_2|\times|V_1|}X_{truth})}
\label{eqn:acc}
\end{equation}
The \emph{objective score} of an algorithm is defined as the objective value obtained corresponding to the consistent matches obtained by that algorithm:
\begin{equation}
obj=J_{gm}(\mathbf{X}_{algthm})
\label{eqn:obj}
\end{equation}
The normalized objective score is the objective scores scaled in $[0,1]$.

The fourth experiment is more focused on an application of inexact or error tolerant subgraph matching. Here we convert the \emph{symbol spotting} problem in graphical documents to a subgraph matching problem and apply the proposed product graph based subgraph matching algorithm for spotting symbols. The performance of the proposed algorithm in presence of a large number of outliers is tested.

All the experiments were done on a workstation with Intel i5 2.50GHz processor and 8GB of RAM\footnote{Our unoptimized Matlab code that we had used for the experimentations is available in \tt{\url{https://sites.google.com/site/2adutta/research}}.}.

\begin{figure}[!h]
\begin{center}
\subfloat{\includegraphics[width=0.95\textwidth]{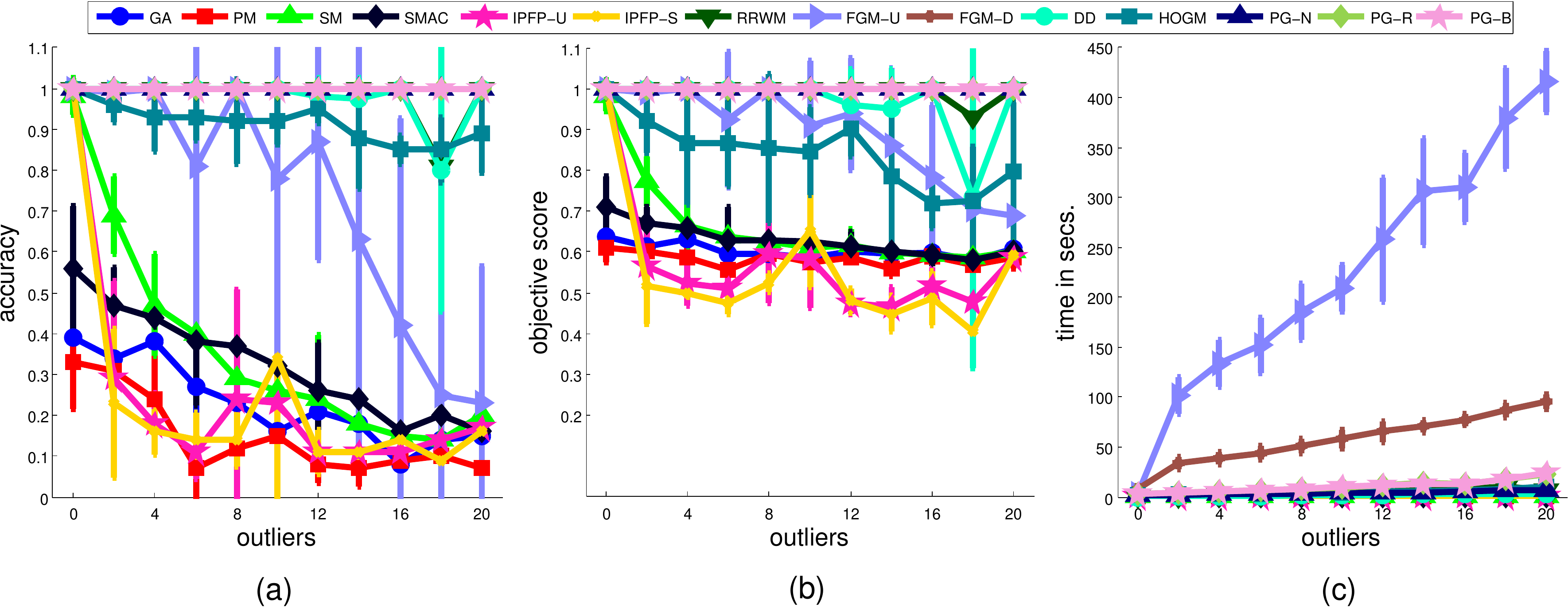}}\\%
\subfloat{\includegraphics[width=0.95\textwidth]{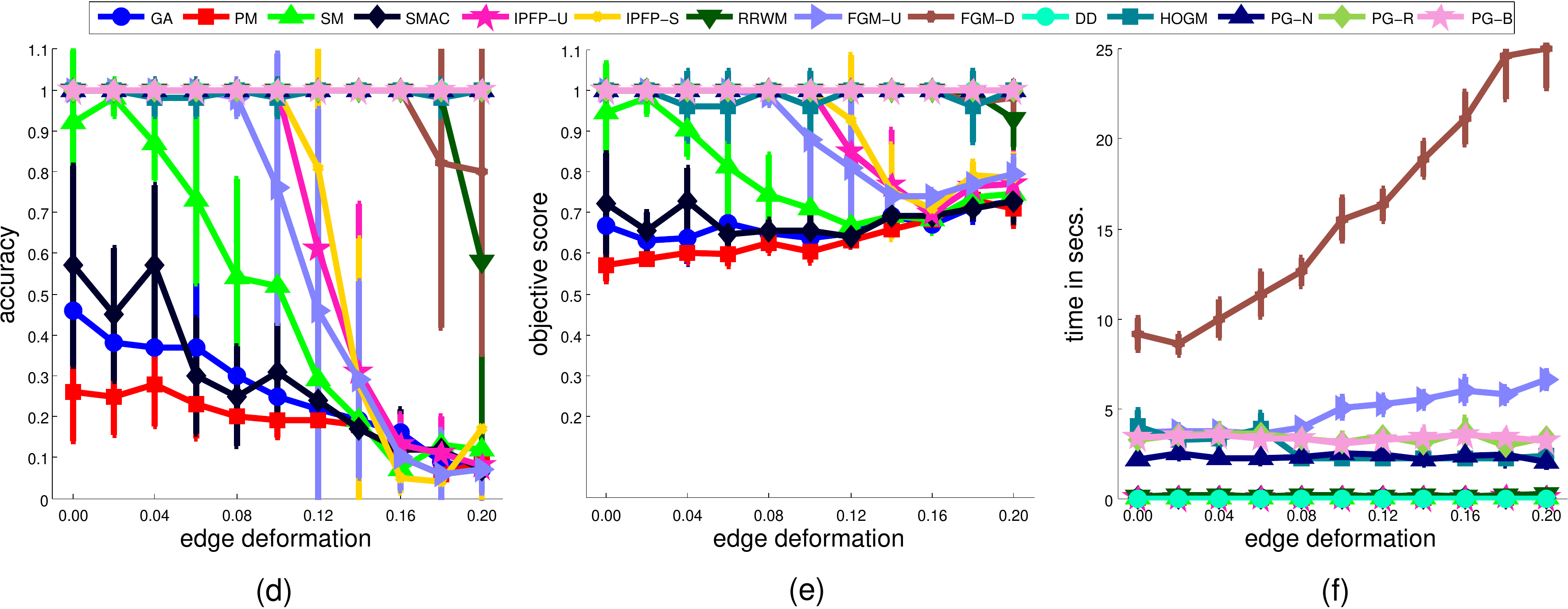}}\\%
\subfloat{\includegraphics[width=0.95\textwidth]{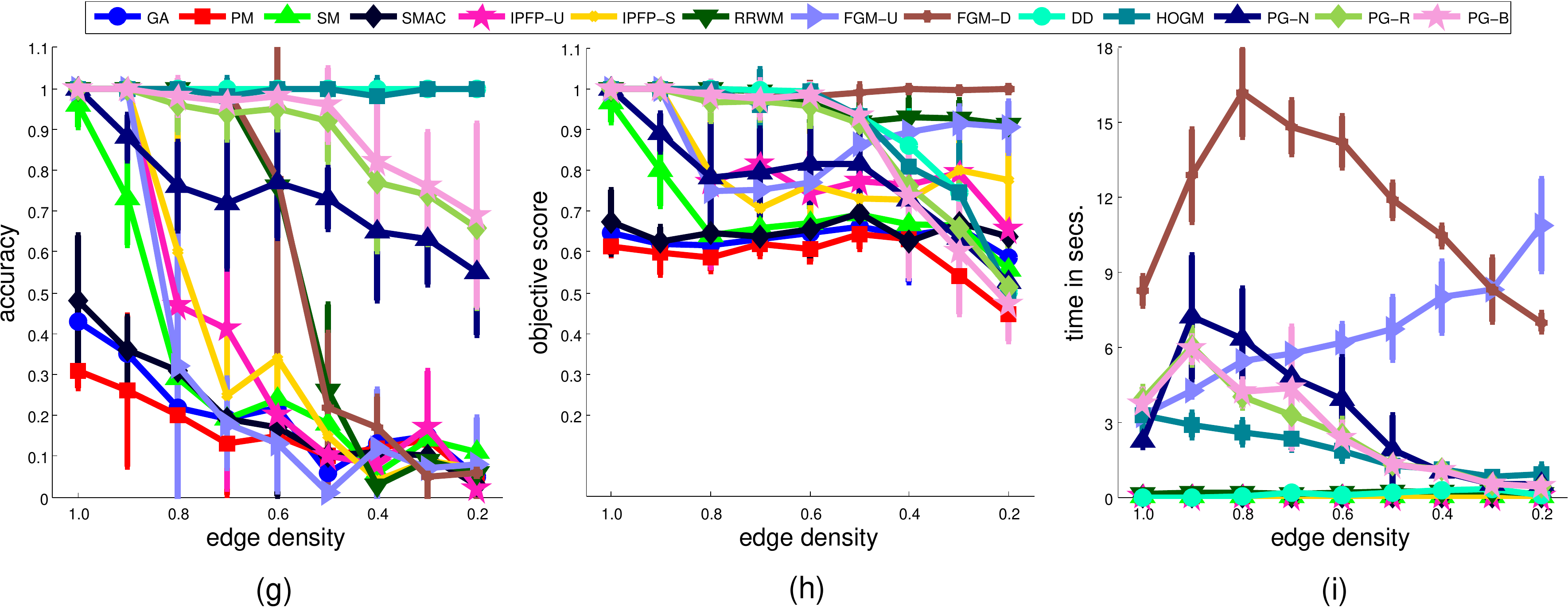}}%
\end{center}
\caption{Comparison of graph matching methods on synthetic datasets: (a) accuracy, (b) objective score and (c) time for outliers experiment; (d) accuracy, (e) objective score and (f) time for edge deformation experiment; (g) accuracy, (h) objective ratio and (i) time for edge density experiment. (Best viewed in pdf.)}
\label{fig:plots-synthetic}
\end{figure}

\subsection{Synthetic dataset}
This particular experiment performed a comparative evaluation of different graph matching algorithms on randomly generated synthetic graphs following the experimental framework shown in~\cite{Gold1996}. This is a common experimental framework and allows one to test graph matching algorithms under various parameter settings. For that, in each run, we constructed two different graphs $G_1(V_1,E_1,\alpha_1,\beta_1)$ and $G_2(V_2,E_2,\alpha_2,\beta_2)$ each of which contained 20 inlier nodes. In each run, we added $n_{out}$ number of outlier nodes to both graphs. Each node $u_i\in V_1$ was assigned with a uniformly distributed node label $\alpha_1(u_i)\backsim\mathcal{U}(0,1)$ and the corresponding node $u_j\in V_2$ was disturbed according to $\alpha_2(u_j) = \alpha_1(u_i)+\epsilon$ by adding a Gaussian noise $\epsilon\backsim\mathcal{N}(0,\sigma^2)$. For each pair of nodes, an edge was randomly created depending on the edge density parameter $\rho\in[0,1]$. Each edge $(u_i,v_i)\in E_1$ was assigned with a uniformly distributed edge label $\beta_1(u_i,v_i)\backsim\mathcal{U}(0,1)$ and the corresponding edge $(u_j,v_j)\in E_2$ was disturbed as $\beta_2(u_j,v_j) = \beta_1(u_i,v_i) + \epsilon$ by adding a Gaussian noise $\epsilon\backsim\mathcal{N}(0,\sigma^2)$. Here it is to be noted that this kind of edge labelling created a directed graph. The node affinity matrix $p_X$ was computed as $p_X=\exp(-\frac{(\alpha_1(u_i)-\alpha_2(u_j))^2}{0.15})$ and the edge affinity matrix $W_X$ was computed as $W_X=\exp(-\frac{(\beta_1(u_i,v_i)-\beta_2(u_j,v_j))^2}{0.15})$.

This experiment tested the performance of graph matching algorithms under three different parameter settings. For each setting, we generated 100 different pairs of $G_1$ and $G_2$ which resulted in 100 independent runs and evaluated the average \emph{accuracy} (\eq{eqn:acc}) and \emph{normalized objective score} (\eq{eqn:obj}) over these 100 runs. In the first setting, we varied the number of outliers ($n_{out}$) 0:2:20 and fixed $\rho=1$ and $\sigma=0$. In the second parameter setting, we varied the noise parameter ($\sigma$) 0:0.02:0.2 and fixed $n_{out}=0$ and $\rho=1$. In the last setting, we varied the edge density parameter ($\rho$) 1.0:-0.1:0.2 and fixed $n_{out}=0$ and $\sigma=0$. In the first two parameter settings, our proposed methods, PG-N, PG-R and PG-B, achieved the best performance together with RRWM, FGM-D in terms of accuracy and objective ratio (see~\fig{fig:plots-synthetic}(a),(b),(d),(e)); although DD and HOGM performed quite closely to the previously mentioned algorithms in these two parameter settings as well. In the last parameter setting, together with most of the other methods, the performance of our algorithms deteriorates as the edge density decreases. This is an obvious phenomenon as any decrement of the edge density destroys the structure of graphs and hence loses the representation power. Our contextual similarities are obtained by propagating the pairwise similarities through the graph edges, which lose discrimination in the absence of edges. This explains the worse performance of our algorithm in this parameter settings. Despite having sparse graphs, DD and HOGM performed quite well as these two algorithms use node affinity, their local distribution in 2-dimensional space and higher order structures as features for matching. In general, the time to find a solution of all the considered algorithms usually increases as the number of outliers and the amount of deformation increases. Nevertheless, the execution time of different algorithms behaves differently as the edge sparsity increases. For example, HOGM performed more efficiently with a decreasing edge density whereas FGM-U performed completely opposite to HOGM. Our algorithms and FGM-D showed slightly different time efficiency. Initially, with the increment of edge sparsity, their time duration increases to get an effective solution. After a certain point it decreases and also the accuracy drops accordingly.
%Among the state-of-the-art algorithms, FGM-D and RRWM performed quite close to ours except in the edge sparseness experiment. Our methods performed exceptionally well due to the fact that we modelled the optimization as node and edge selection problem of the product graph which does not get affected by the absence of edges.

\begin{figure}[!h]
\begin{center}
\subfloat{\includegraphics[width=0.7\textwidth]{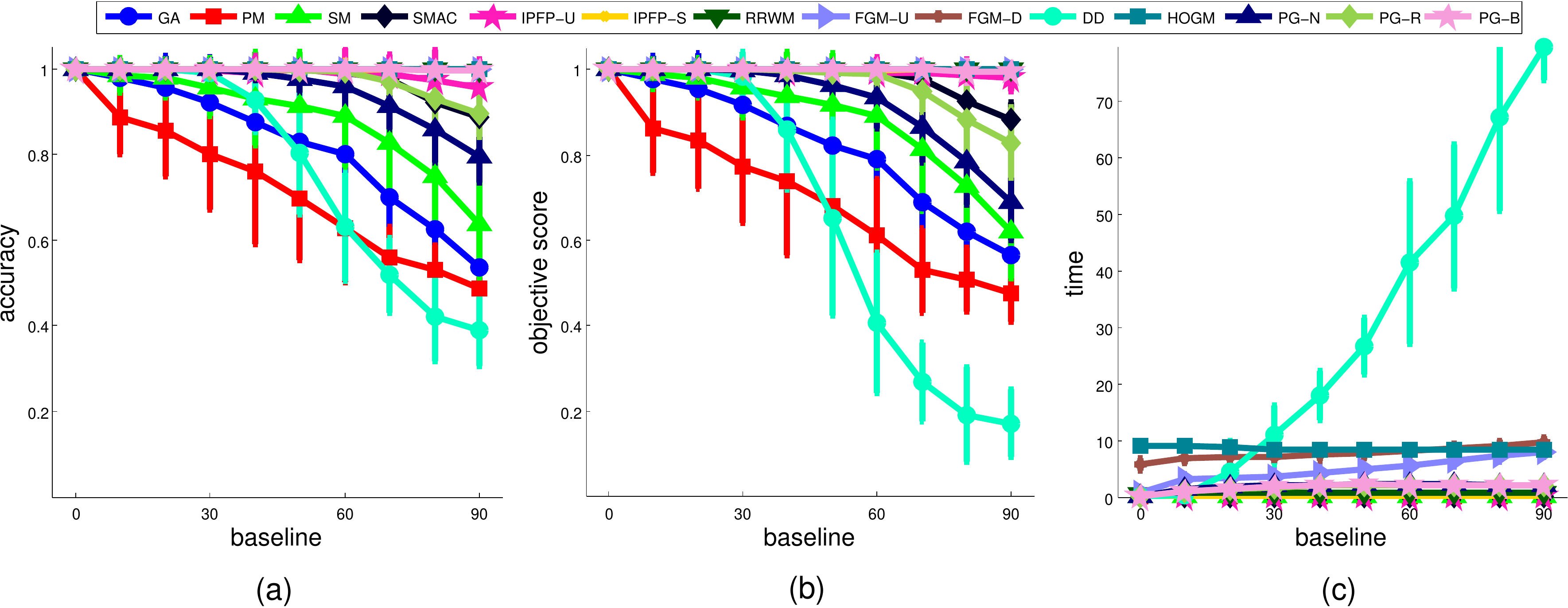}}\\
%\hspace{0.8mm}
\subfloat{\includegraphics[width=0.7\textwidth]{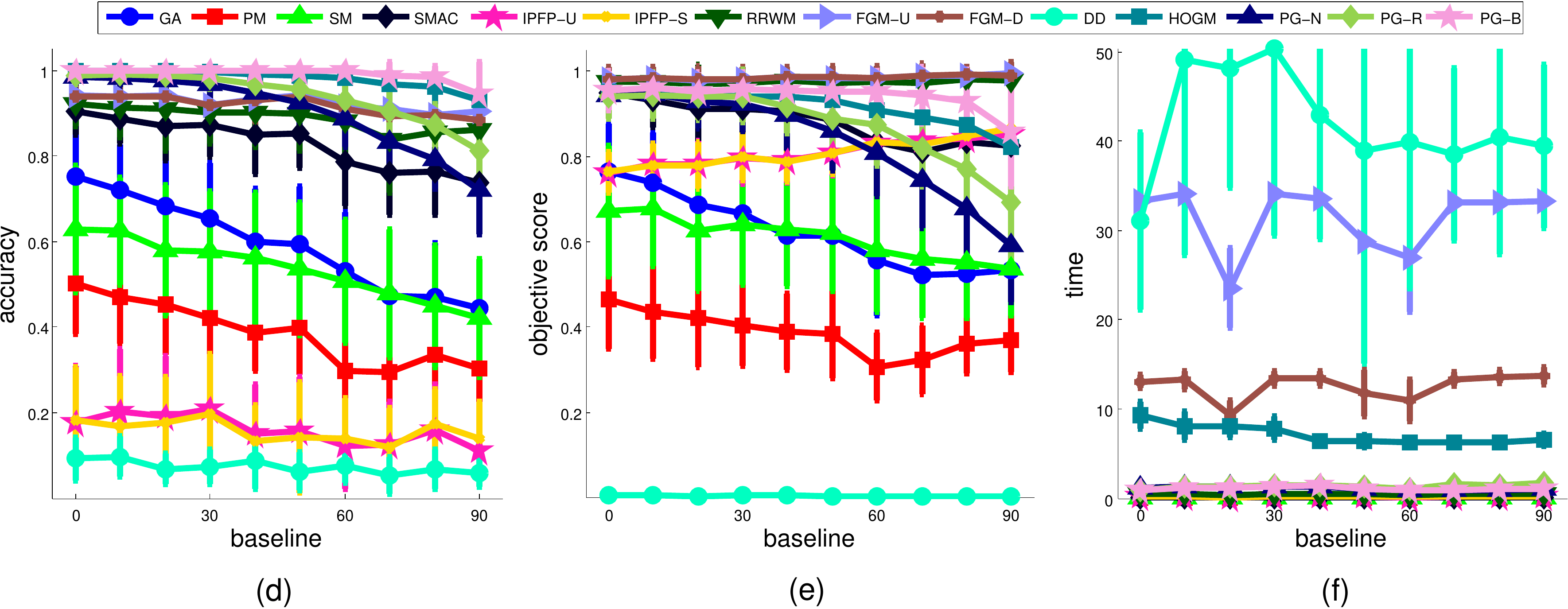}}\\%
\subfloat{\includegraphics[width=0.7\textwidth]{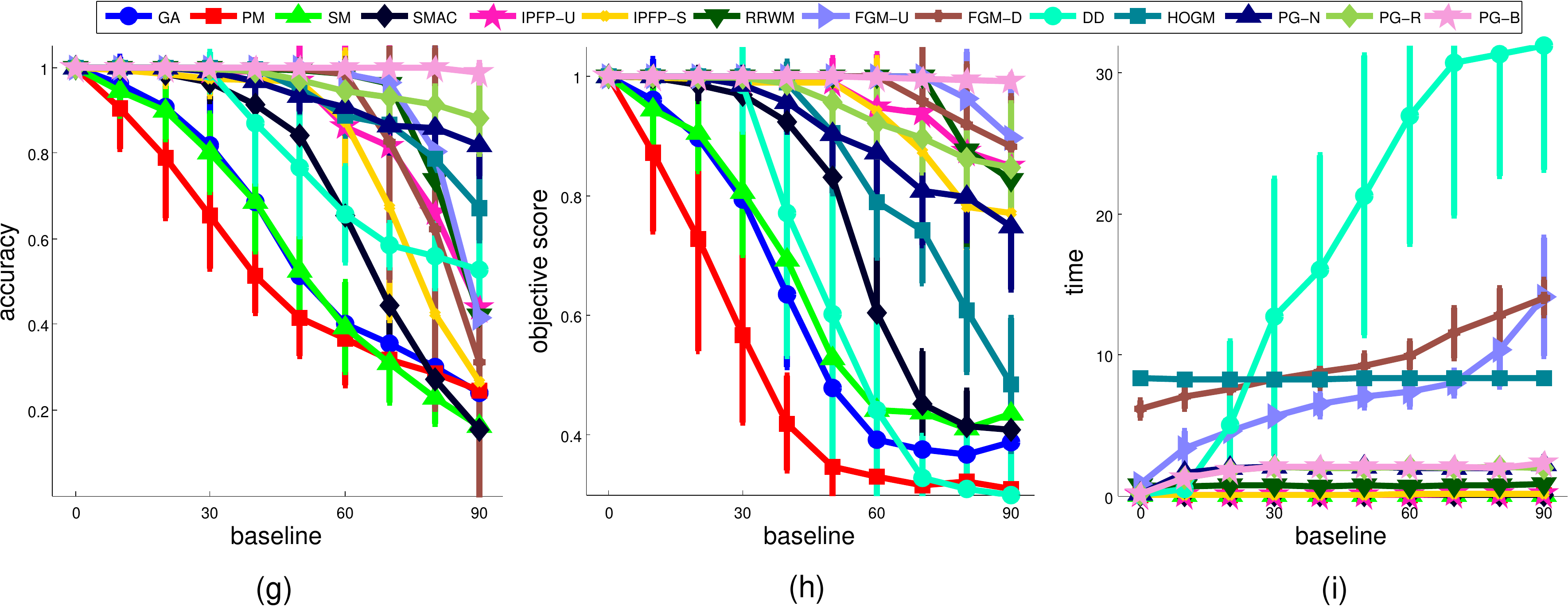}}\\
%\hspace{0.8mm
\subfloat{\includegraphics[width=0.7\textwidth]{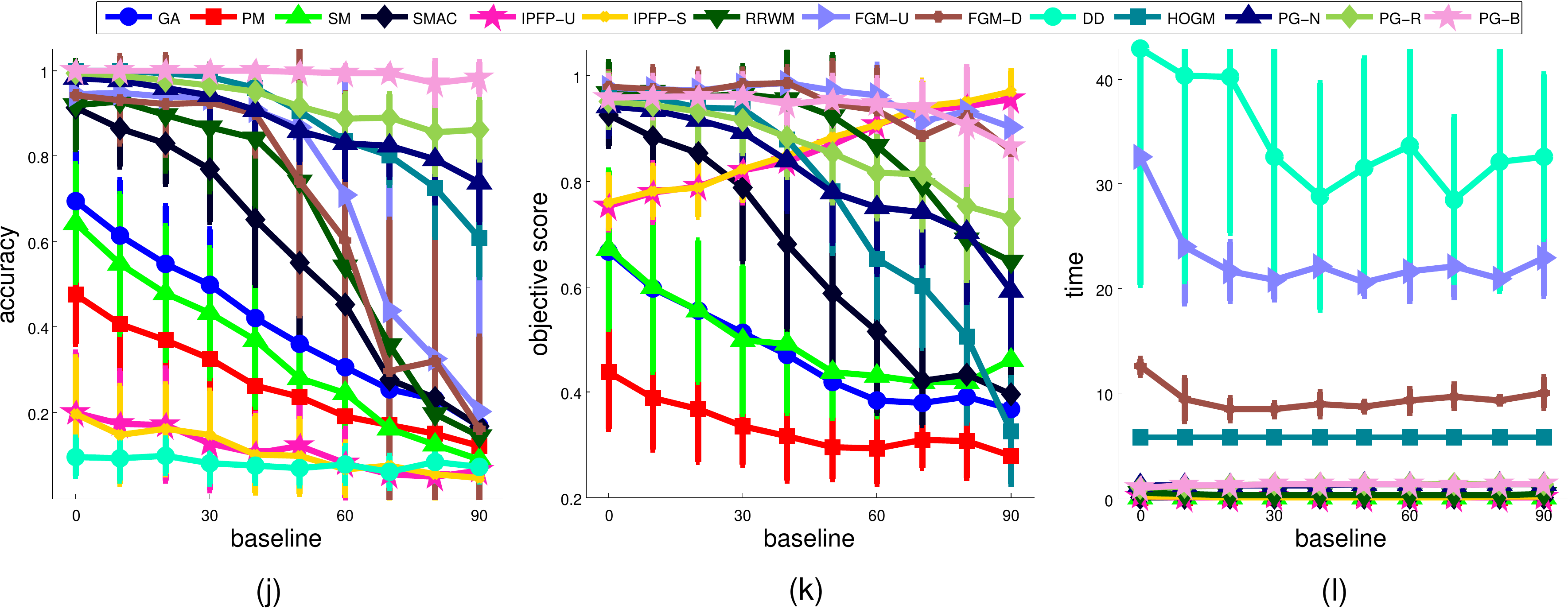}}
\end{center}
\caption{Comparison of graph matching methods in CMU house and hotel datasets: (a) accuracy, (b) objective score and (c) time on house image sequence using 30 nodes; (d) accuracy, (e) objective score and (f) time on house image dataset using 25 nodes; (g) accuracy, (h) objective score and (i) time on hotel image sequence using 30 nodes, (j) accuracy, (k) objective ratio and (l) time on hotel image dataset using 25 nodes. (Best viewed in pdf.)}
\label{fig:plots-cmu}
\end{figure}

\subsection{House and Hotel image dataset}
\label{ssec:expt-cmu}
The CMU house and hotel image sequences\footnote{\tt{\url{http://vasc.ri.cmu.edu//idb/html/motion}}} are among the most popular datasets to test the performance of graph matching algorithms~\cite{Gold1996,Caetano2009,Cho2010,Leordeanu2012,Zhou2012}. This dataset provides a more realistic way to judge the performance of graph matching algorithms. It consists of 111 frames of a toy house and 101 frames of a toy hotel, each of which had been manually labelled with 30 landmark points. Each individual frame in these sequences has a certain difference in the position of viewpoint with respect to its adjacent frames. We used Delaunay triangulation to connect the landmarks. Each node was assigned with a shape context descriptor. The edge label $\beta$ was computed as the length of the edge, here it is to be mentioned that this kind of edge labelling created undirected edges. In this experiment both the node and edge affinities were computed as dot product similarities. We performed the experiment with all possible image pairs, spaced by 0:10:90 frames. To evaluate the performance we computed the average accuracy (\eq{eqn:acc}) and normalized objective score (\eq{eqn:obj}) in each sequence gap.

\begin{figure}[!ht]
\subfloat[]{\label{sfig:house-matchings}\includegraphics[width=0.49\textwidth,height=2.6cm]{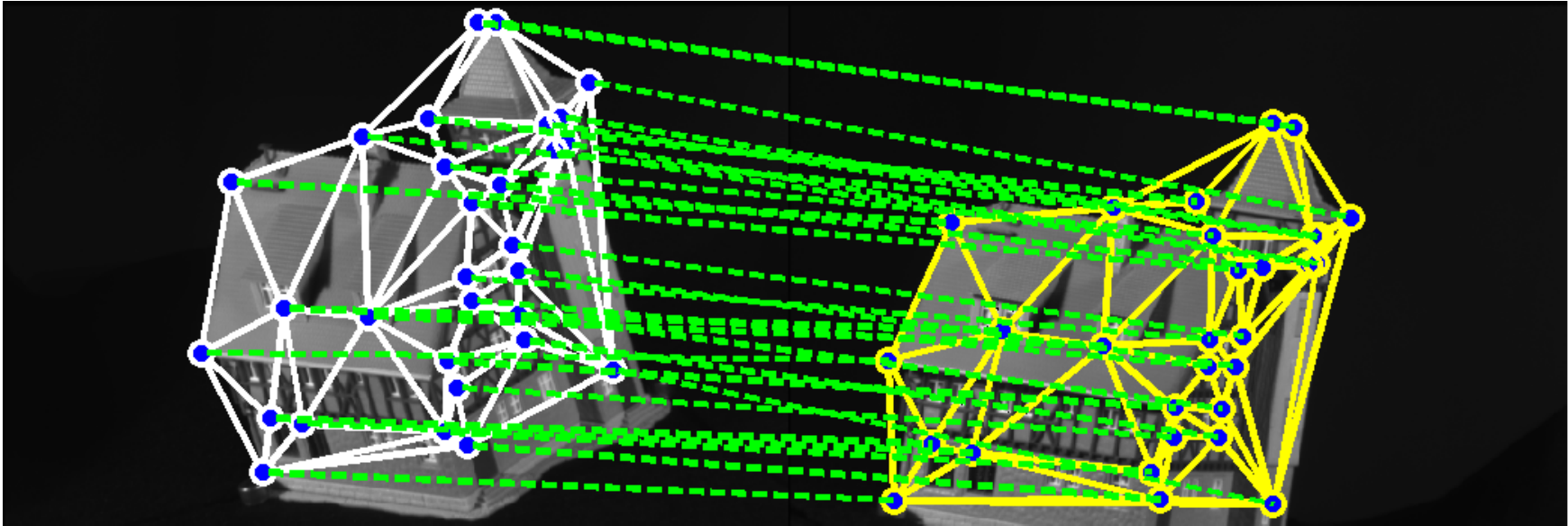}}
\hspace{0.5mm}
\subfloat[]{\label{sfig:hotel-matchings}\includegraphics[width=0.49\textwidth,height=2.6cm]{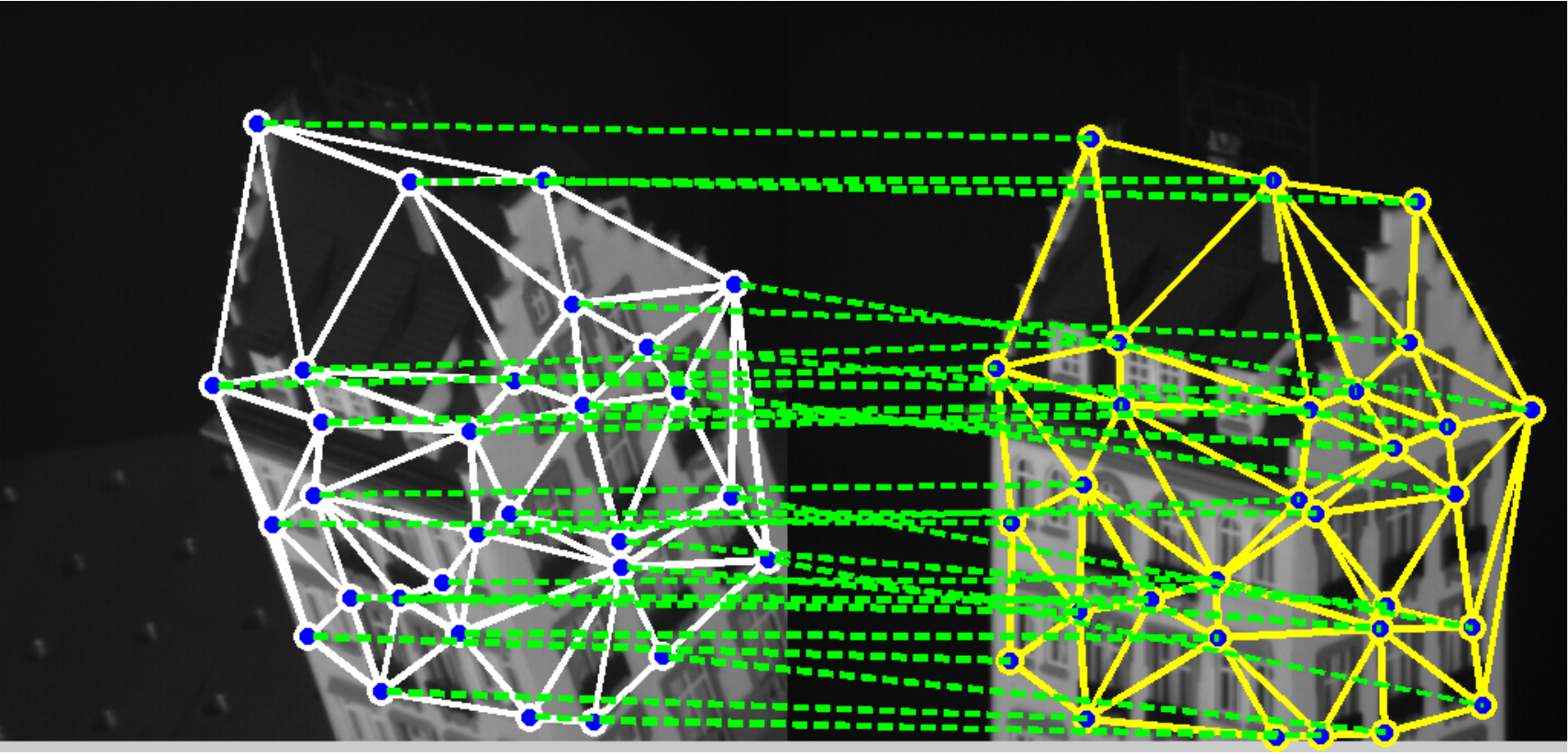}}
\caption{Qualitative results obtained by our method (PG-B): (a) for house sequence with baseline difference 90 (b) for hotel sequence with baseline difference 60. (Best viewed in pdf.)}
\label{fig:qual-cmu}
\end{figure}

For both sequences, we tested the performance of graph matching methods under two scenarios. In the first case, we used all the 30 nodes (annotated landmark points) and in the second one we randomly chose 25 nodes from both the graphs and performed a subgraph matching. It can be seen that for the house sequence in the first case, IPFP-U, IPFP-S, RRWM, FGM-U, FGM-D, HOGM and PG-B obtained almost perfect matchings of the original graphs (see \fig{fig:plots-cmu}(a)-(b)). But for the second case, as some of the nodes from both the graphs got hidden and graphs got corrupted, performance of all the methods deteriorated (see \fig{fig:plots-cmu}(d)-(e)). Still our method PG-B performed best among all. Here the difference in performance between the method working with backtrackless walk (PG-B) and the method working with random walk (PG-R) is to be noticed. Since the graphs considered for this experiment are undirected, removal of tottering proves to be effective. Nevertheless, the difference in performance between the method working with pairwise similarities (PG-N) and the methods working with contextual similarities (PG-R, PG-B) is to be noticed. Since both PG-R and PG-B are taking into account contextual information they performed better than PG-N. The hotel sequence appeared as a harder challenge to all the algorithms, as most of them performed worse than the house sequence. Here all the methods performed worse with distorted graphs (see \fig{fig:plots-cmu}(j)-(k)) than their undistorted versions (see \fig{fig:plots-cmu}(g)-(h)). Here also in both cases our methods with contextual similarities performed best. Qualitative results obtained by PG-B on this dataset are shown in~\fig{fig:qual-cmu}. In case of an undistorted pair of graphs, the execution time of most of the methods increases with the baseline. In difficult cases such as increased baseline, DD takes a lot of iterations to converge. This explains its significant increase of execution time as well as the poor results with the increment of baseline.

\subsection{Car and Motorbike image dataset}
The car and motorbike image dataset\footnote{\tt{\url{https://goo.gl/CCp9lA}}} was created in~\cite{Leordeanu2012}. This dataset consists of 30 pairs of car images and 20 pairs of motorbike images that were taken from the dataset of PASCAL challenges. Each pair of images consists of $15\backsim 52$ true correspondences. Since these graphs contain a rather large number of outlier nodes, this experiment provides a harder challenge. The node label $\alpha$ was computed as the orientation of the normal vector to each of the contours. Like before, we adopted Delaunay triangulation to build the graph. For this experiment, edge label $\beta$ was computed as the edge lengths and angles with the horizontal. Due to the nature of the edge labels, the graphs constructed were directed.

\begin{figure}[!ht]
\subfloat[]{\label{sfig:car-matchings}\includegraphics[width=0.49\textwidth,height=2.6cm]{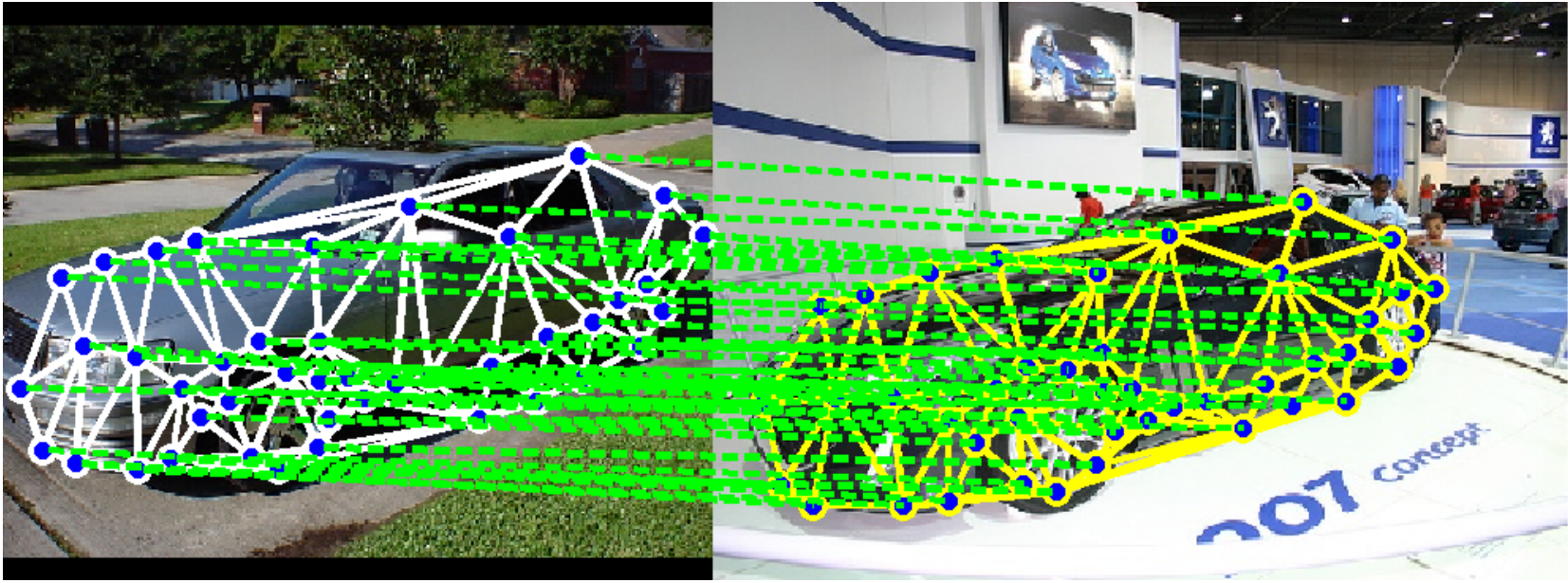}}
\hspace{0.5mm}
\subfloat[]{\label{sfig:motorbike-matchings}\includegraphics[width=0.49\textwidth,height=2.6cm]{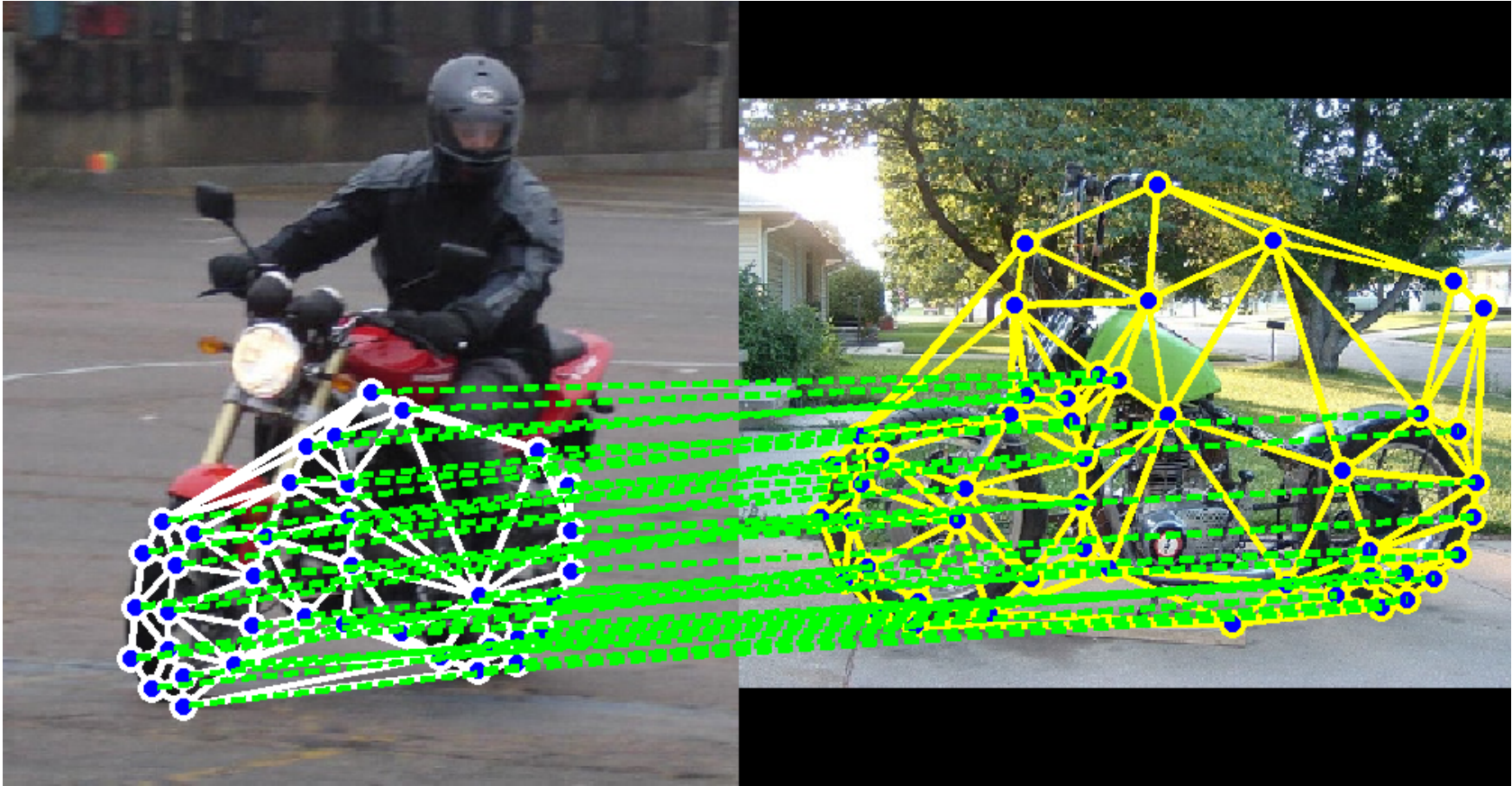}}
\caption{Qualitative results obtained by our method (PG-B): (a) for car sequence with no outliers, (b) for motorbike sequence with outliers. (Best viewed in pdf.)}
\label{fig:qual-pascal}
\end{figure}

\begin{figure}[!h]
\subfloat{\includegraphics[width=\textwidth]{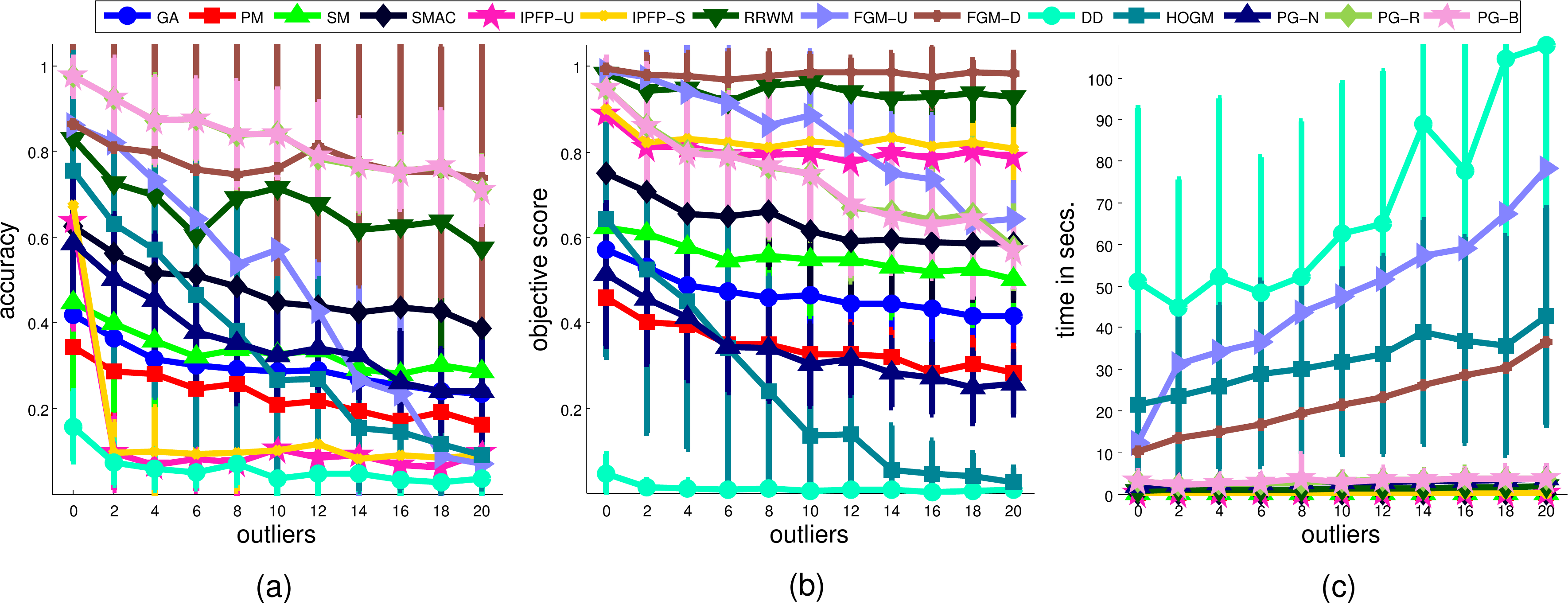}}\\
%\hspace{0.8mm}
\subfloat{\includegraphics[width=\textwidth]{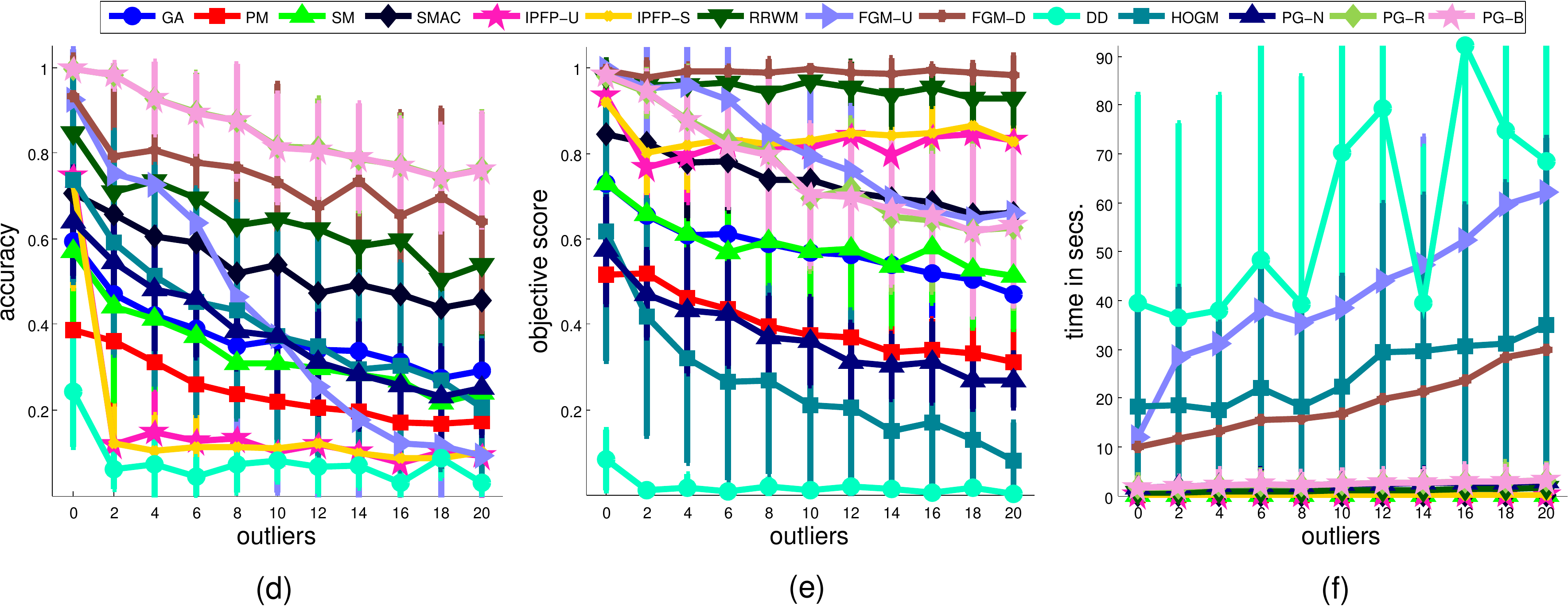}}\\
\caption{Comparison of graph matching methods on car and motorbike datasets: (a) accuracy, (b) objective score and (c) time on car images; (d) accuracy, (e) objective score and (f) time on motorbike images. (Best viewed in pdf.)}
\label{fig:plots-pascal}
\end{figure}

We tested the performance of the graph matching methods against noise and for that we added $0:2:20$ outlier nodes to both graphs. For performance evaluation we computed the average accuracy (\eq{eqn:acc}) and normalized objective score (\eq{eqn:obj}) within a level of outliers (see~\fig{fig:plots-pascal}). The performance of the algorithms is quite similar. Particularly, FGM-D, PG-R and PG-B were quite competitive and obtained similar performance. In both cases, PG-N performed worse than PG-R and PG-B due to the same reason as explained previously. As in this case the graphs considered were directed, very minor difference in performance was observed between PG-R and PG-B. Qualitative results obtained by PG-B on this dataset are shown in~\fig{fig:qual-pascal}.

\subsection{Symbol spotting as an inexact subgraph matching problem}
\label{ssec:expt-ss}
\begin{figure*}[t]
\begin{minipage}{0.25\textwidth}
\begin{center}
\subfloat[]{\label{sfig:example-fps}\rotatebox{180}{\reflectbox{\includegraphics[width=\textwidth]{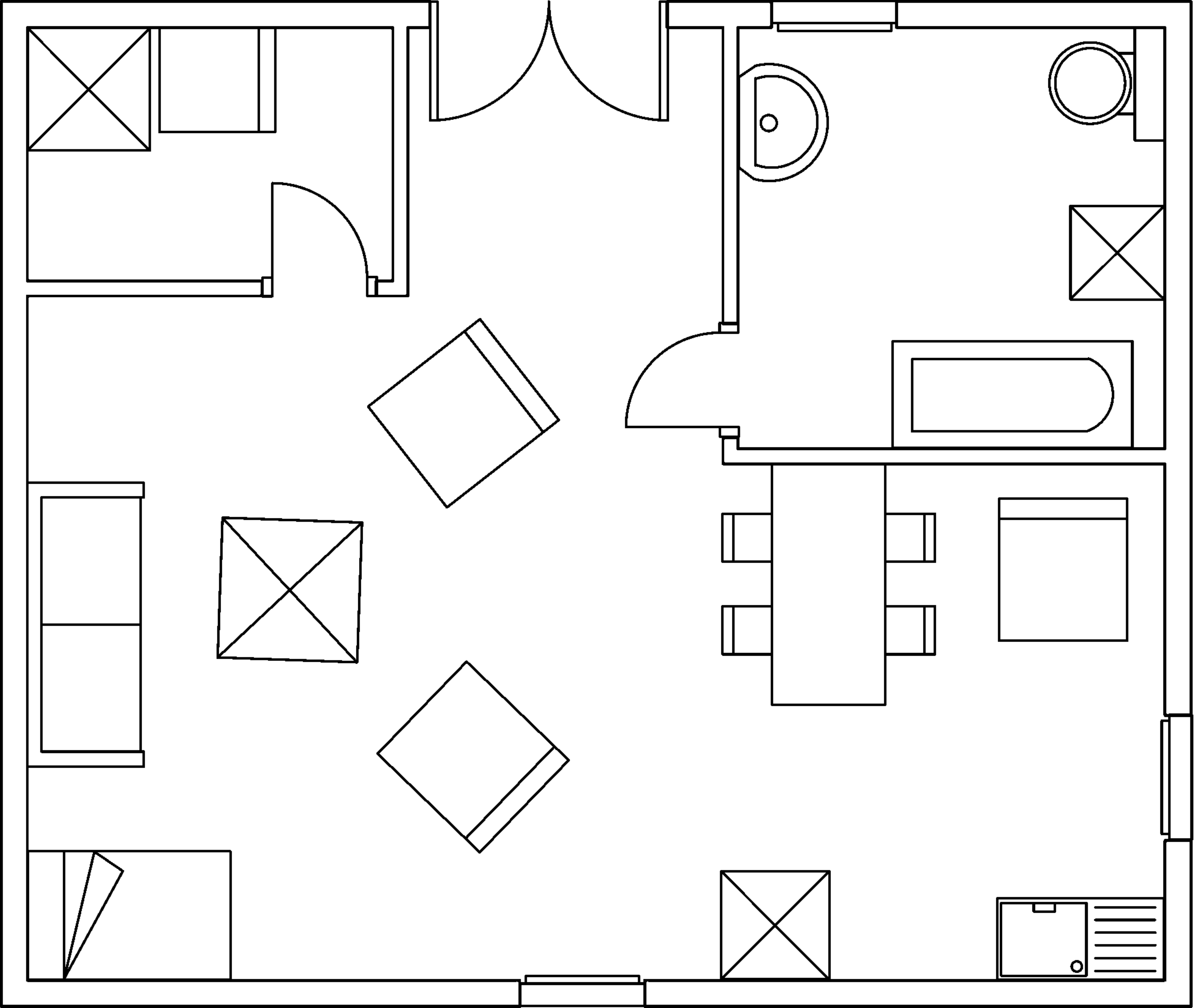}}}}
\end{center}
\end{minipage}
\begin{minipage}[t]{0.75\textwidth}
\begin{center}
\subfloat[]{\label{sfig:example-armchair}\includegraphics[width=0.08\textwidth]{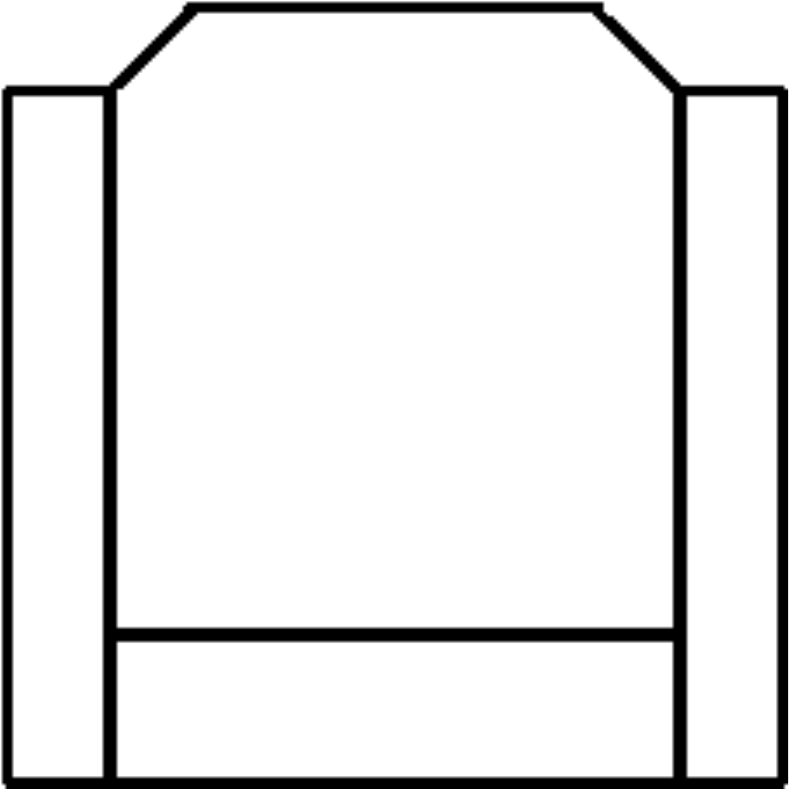}}
\hspace{0.5mm}
\subfloat[]{\label{sfig:example-bed}\includegraphics[width=0.08\textwidth]{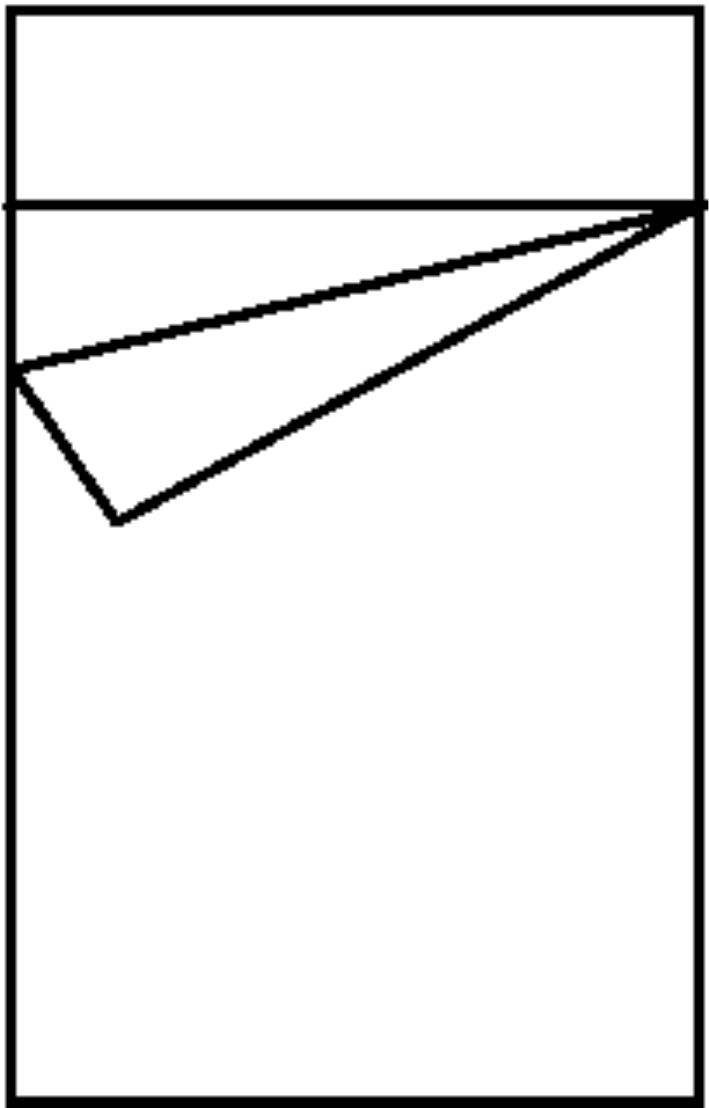}}
\hspace{0.5mm}
\subfloat[]{\label{sfig:example-door1}\includegraphics[width=0.08\textwidth]{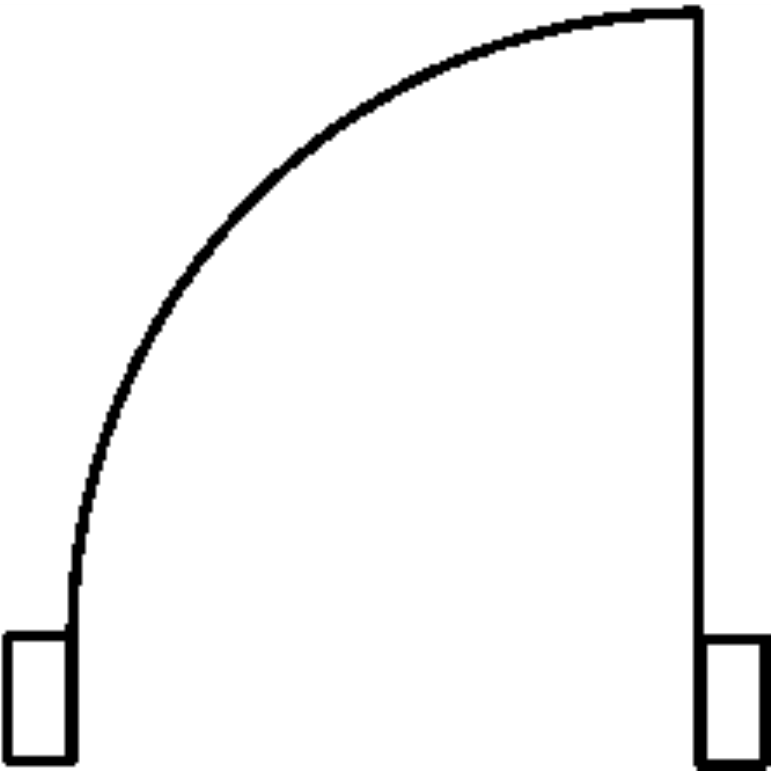}}
\hspace{0.5mm}
\subfloat[]{\label{sfig:example-door2}\includegraphics[width=0.12\textwidth]{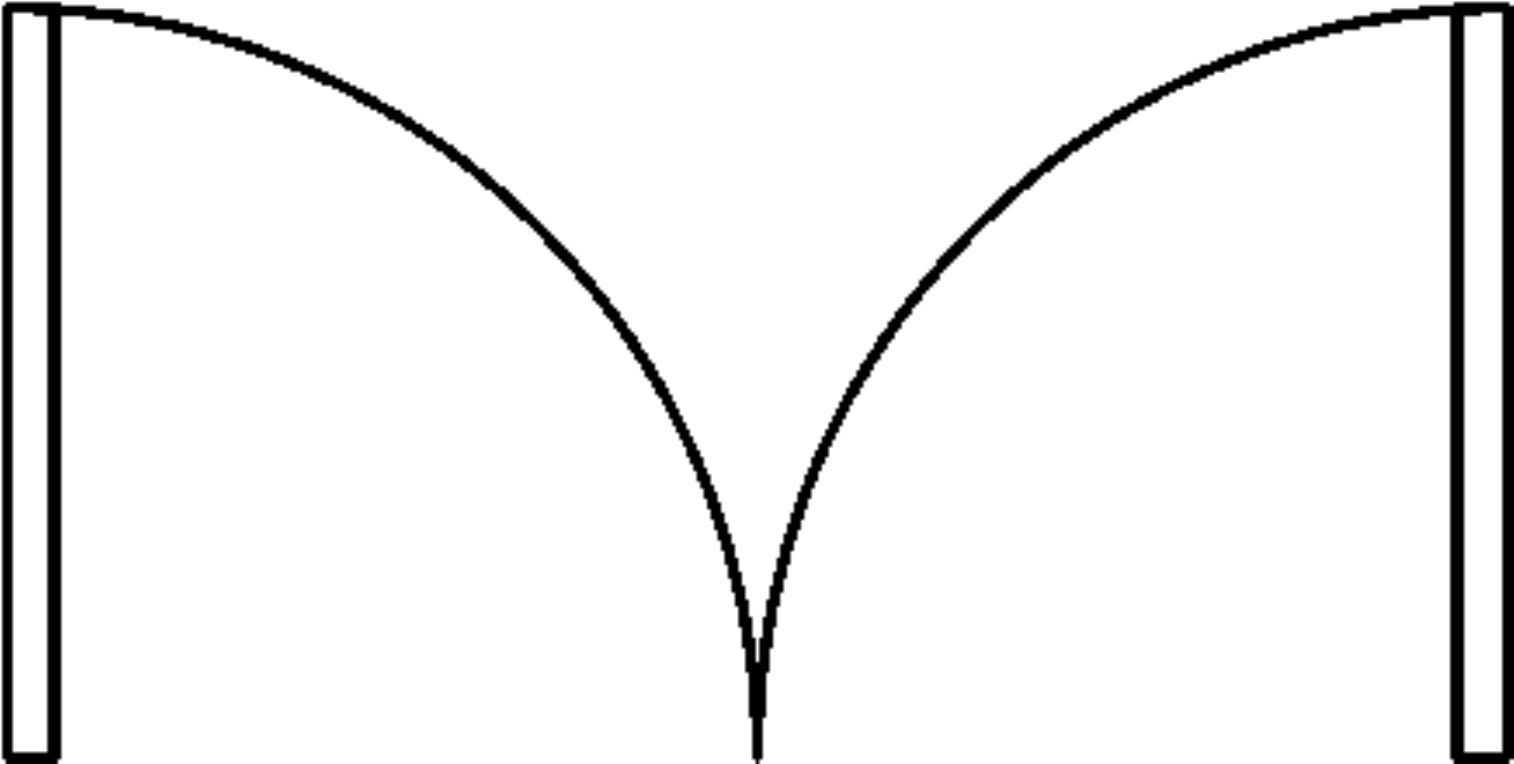}}
\hspace{0.5mm}
\subfloat[]{\label{sfig:example-sink1}\includegraphics[width=0.08\textwidth]{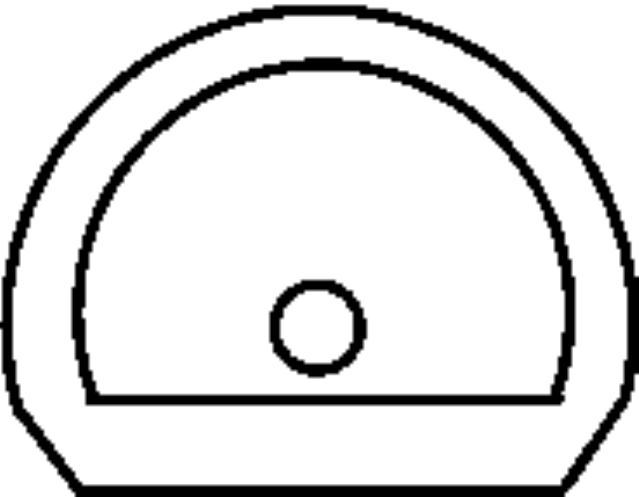}}
\hspace{0.5mm}
\subfloat[]{\label{sfig:example-sink2}\includegraphics[width=0.16\textwidth]{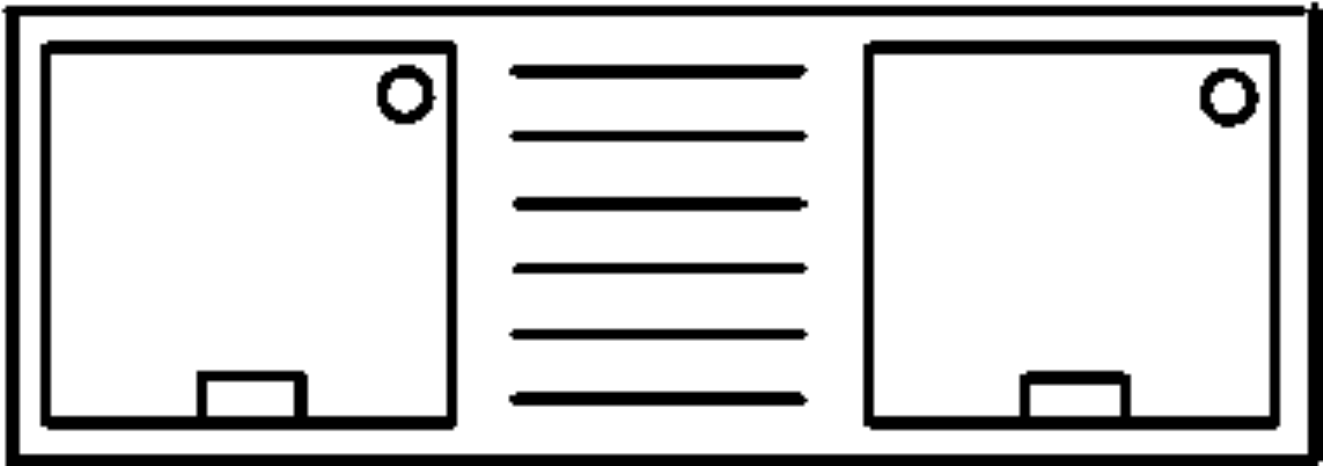}}
\hspace{0.5mm}
\subfloat[]{\label{sfig:example-sink3}\includegraphics[width=0.1\textwidth]{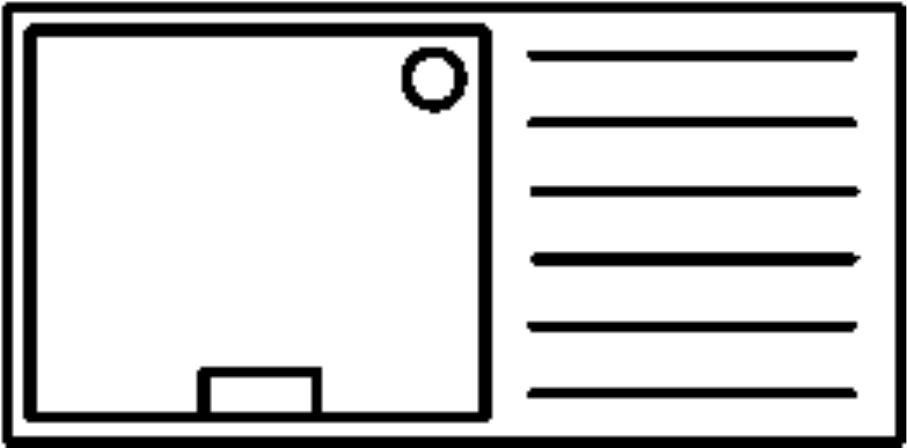}}
\hspace{0.5mm}
\subfloat[]{\label{sfig:example-sink4}\includegraphics[width=0.1\textwidth]{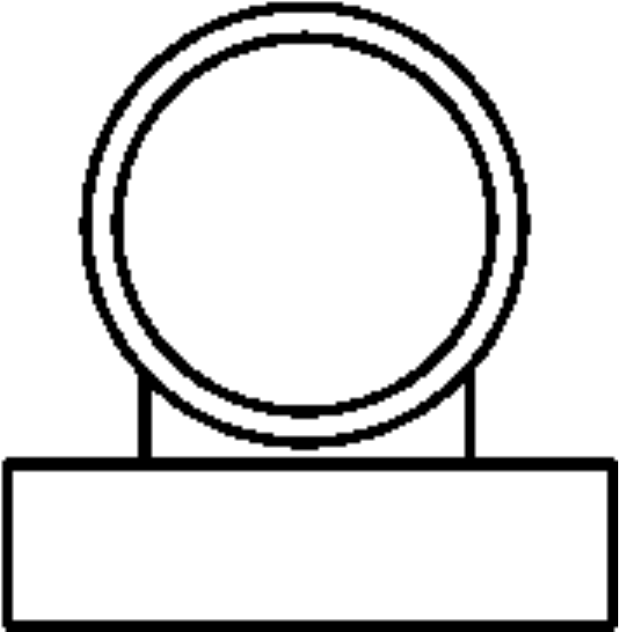}}\\
\hspace{0.5mm}
\subfloat[]{\label{sfig:example-sofa1}\includegraphics[width=0.08\textwidth]{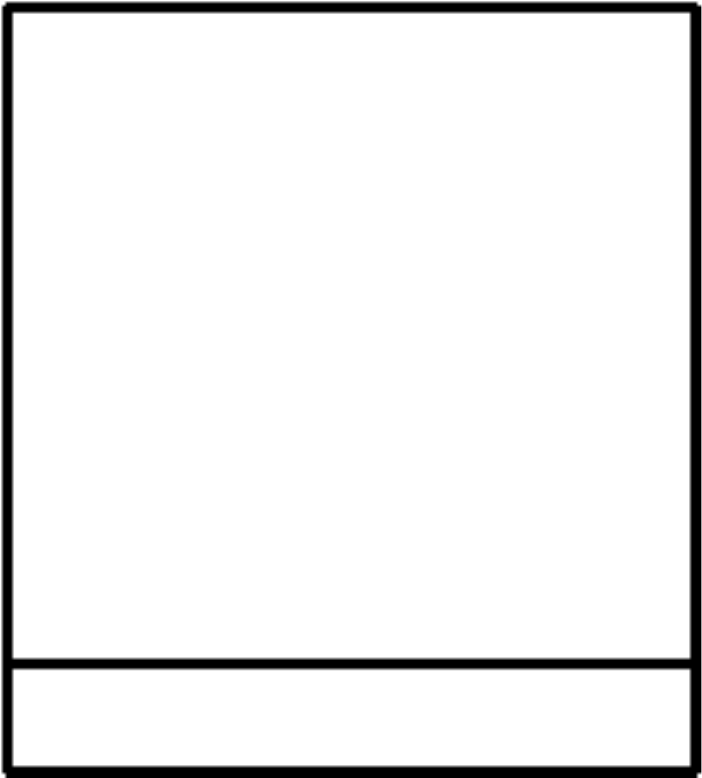}}
\hspace{0.5mm}
\subfloat[]{\label{sfig:example-sofa2}\includegraphics[width=0.16\textwidth]{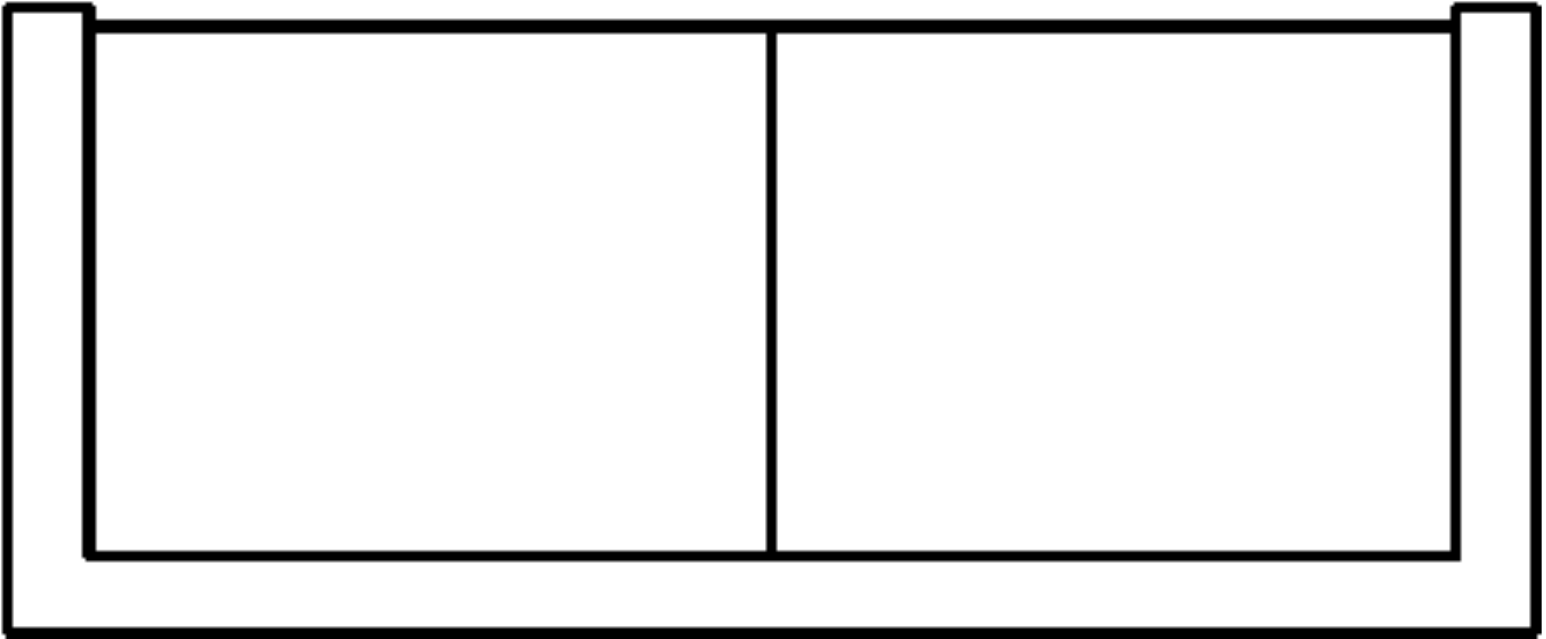}}
\hspace{0.5mm}
\subfloat[]{\label{sfig:example-table1}\includegraphics[width=0.08\textwidth]{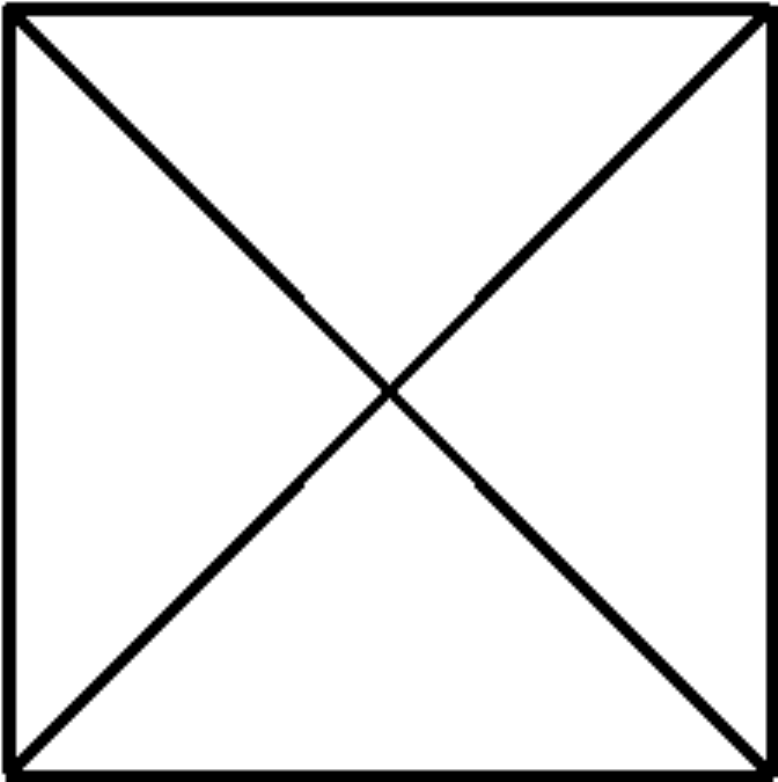}}
\hspace{0.5mm}
\subfloat[]{\label{sfig:example-table2}\includegraphics[width=0.12\textwidth]{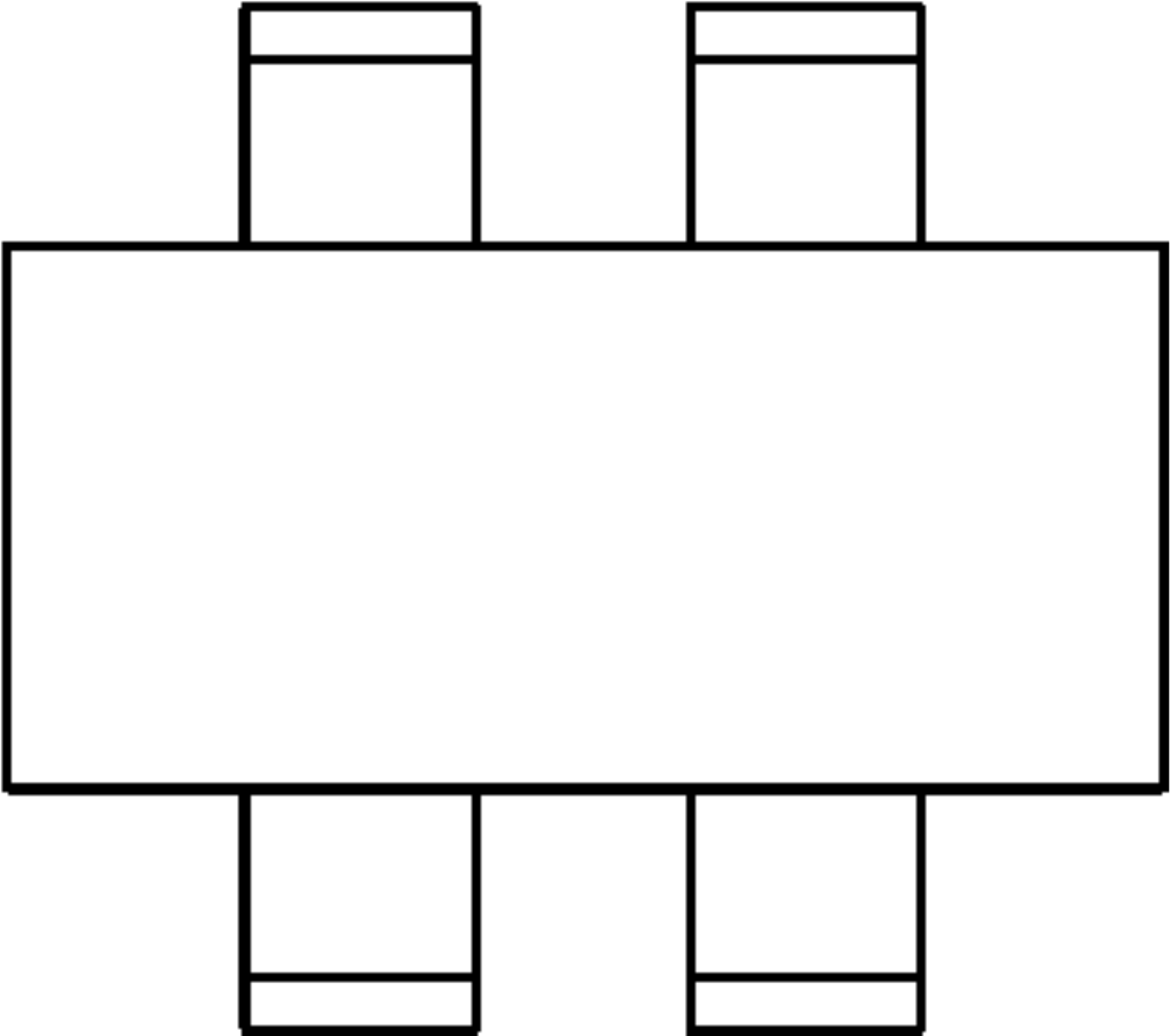}}
\hspace{0.5mm}
\subfloat[]{\label{sfig:example-table3}\includegraphics[width=0.1\textwidth]{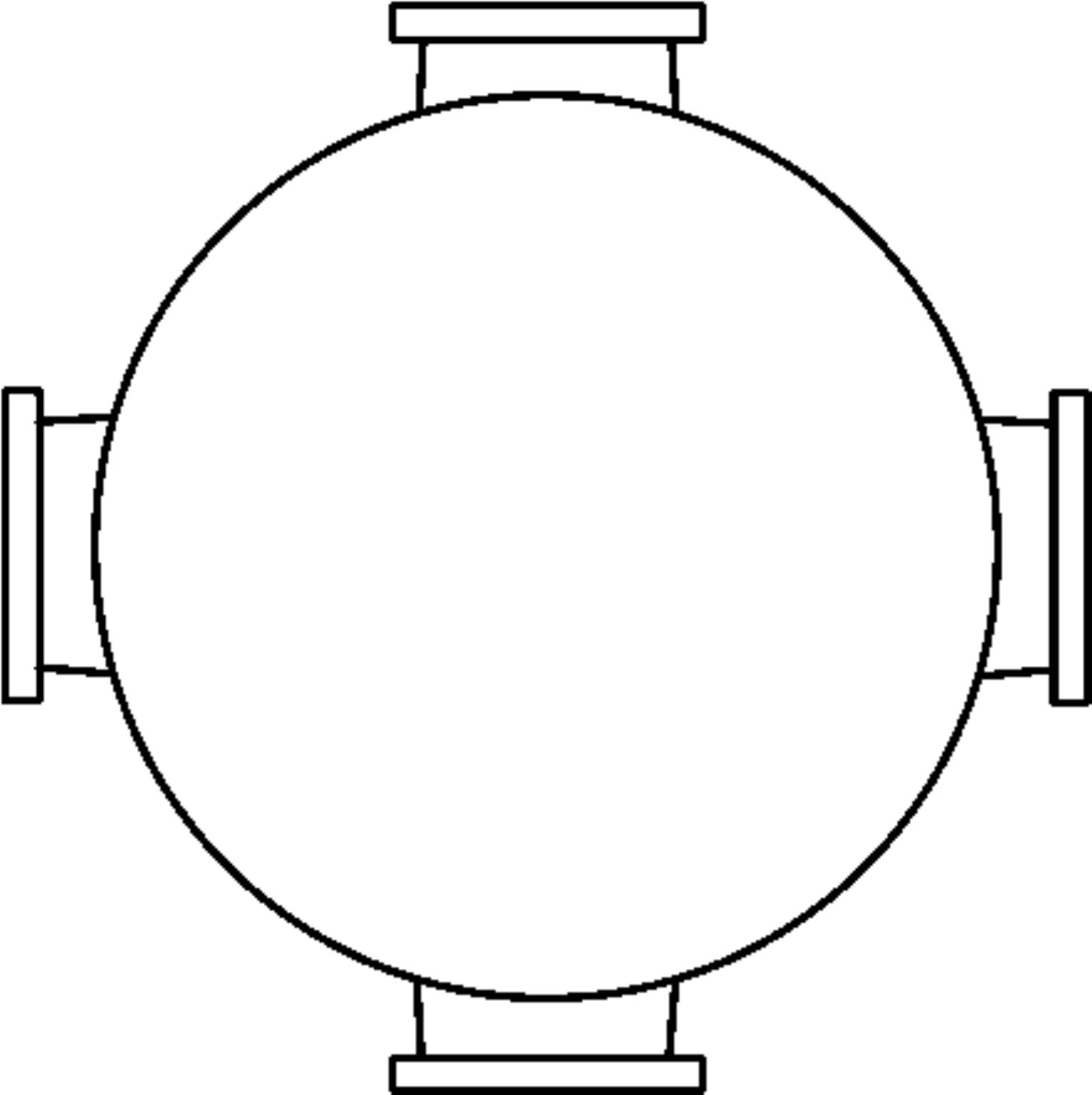}}
\hspace{0.5mm}
\subfloat[]{\label{sfig:example-tub}\includegraphics[width=0.1\textwidth]{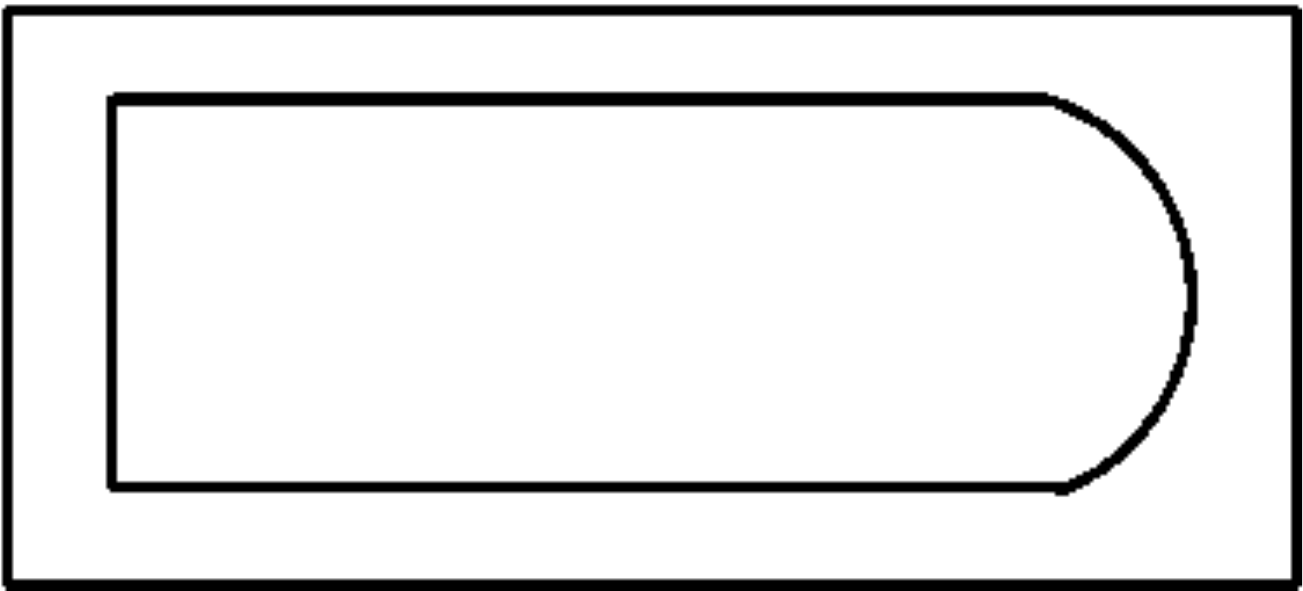}}
\hspace{0.5mm}
\subfloat[]{\label{sfig:example-window1}\includegraphics[width=0.1\textwidth]{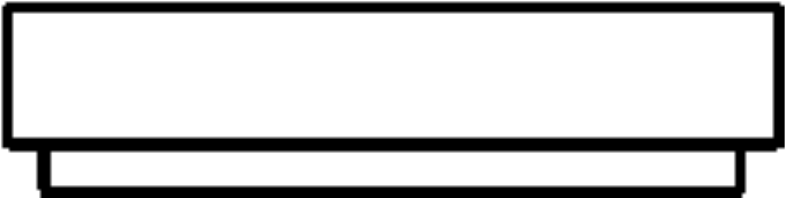}}
\hspace{0.5mm}
\subfloat[]{\label{sfig:example-window2}\includegraphics[width=0.1\textwidth]{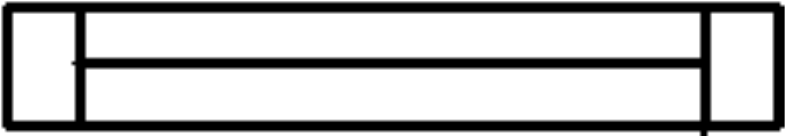}}
\end{center}
\end{minipage}
\caption{(a) Example of a floorplan from SESYD dataset and examples of different isolated symbols from the same repository: (b) \emph{armchair}, (c) \emph{bed}, (d) \emph{door1}, (e) \emph{door2}, (f) \emph{sink1}, (g) \emph{sink2}, (h) \emph{sink3}, (i) \emph{sink4}, (j) \emph{sofa1}, (k) \emph{sofa2}, (l) \emph{table1}, (m) \emph{table2}, (n) \emph{table3}, (o) \emph{tub}, (p) \emph{window1}, (q) \emph{window2}.}
\end{figure*}

Symbol spotting is a well known problem in graphic recognition which is a subfield of document image analysis. It can be defined as locating a given query symbol in large graphical documents such as line drawings, floorplan, circuit diagrams etc. As line drawings have especial structural and syntactic nature, symbol spotting is often devised as subgraph matching problem~\cite{LladosPAMI2001,Wenyin2007,LeBodic2012}. Due to the existence of adequate background information, structural noise, large target graphs, this experiment presents a hard challenge to a subgraph matching algorithm. We considered the SESYD dataset\footnote{\tt{\url{http://mathieu.delalandre.free.fr/projects/sesyd}}}~\cite{Delalandre2008} for this experiment. The original SESYD dataset is a collection of 10 different subsets of floorplan images (shown in \fig{sfig:example-fps}), where each of the sets is created by putting different isolated symbols (see \fig{sfig:example-armchair}-\fig{sfig:example-window2}) in different orientations, scales etc. For this particular experiment we randomly selected 20 images from each of the 10 subsets which in total resulted in 200 images. This experiment provides a complete framework for inexact subgraph matching from application point of view.

For representing these graphical documents we considered the \emph{dual edge graph} representation proposed in~\cite{Dutta2013c}. Let $G_1=(V_1,E_1,\alpha_1,\beta_1)$ and $G_2=(V_2,E_2,\alpha_2,\beta_2)$ be respectively the attributed dual edge graphs for the pattern and the target. The target graphs in this dataset contain far more outlier nodes than any of the datasets described before. Here $\alpha:V_1(u)\rightarrow \mathbb{R}^{m\times 7}$ is a node labelling function and is defined as the set of seven Hu moments invariants~\cite{Hu1962} of the acyclic graph paths joining the extremities of an edge in the edge graph ($m$ is total number of paths). $\beta:E_1(u_i,v_i)\rightarrow \mathbb{R}^3$ is an edge labelling function and is constructed of three components: 

\begin{itemize}
\item normalized angle (angle divided by $180^\circ$) between two edges (of the edge graph) joined by the dual edge.
\item the ratio of the two edges (of the edge graph) joined by the dual edge.
\item the ratio of the distance between the mid points of the two edges (of the edge graph) and the total length of the edges joined by the dual edge.
\end{itemize}
For computing the distance between the dual nodes we used a modified version of Hausdorff distance~\cite{Fischer2013} as follows:
\[
d(A,B)=\sum_{a\in A} \min_{b\in B} dm(a,b) + \sum_{b\in B} \min_{a\in A} dm(a,b)
\]
Here $dm(a,b)$ denotes the Euclidean distance of $a\in A$ from $b\in B$ and $d(A,B)$ denotes the distance between the sets $A$ and $B$. Since the node label in our dual edge graph representation is a set of Hu moments, we used this particular distance measure. Then the affinities were obtained as $\exp{(-d)}$. For having the distance between the edges we used the Euclidean distance and here also their similarity were obtained as $\exp{(-d)}$.

\begin{figure*}[!t]
\begin{center}
\subfloat[]{\includegraphics[width=0.32\textwidth,height=0.24\textwidth]{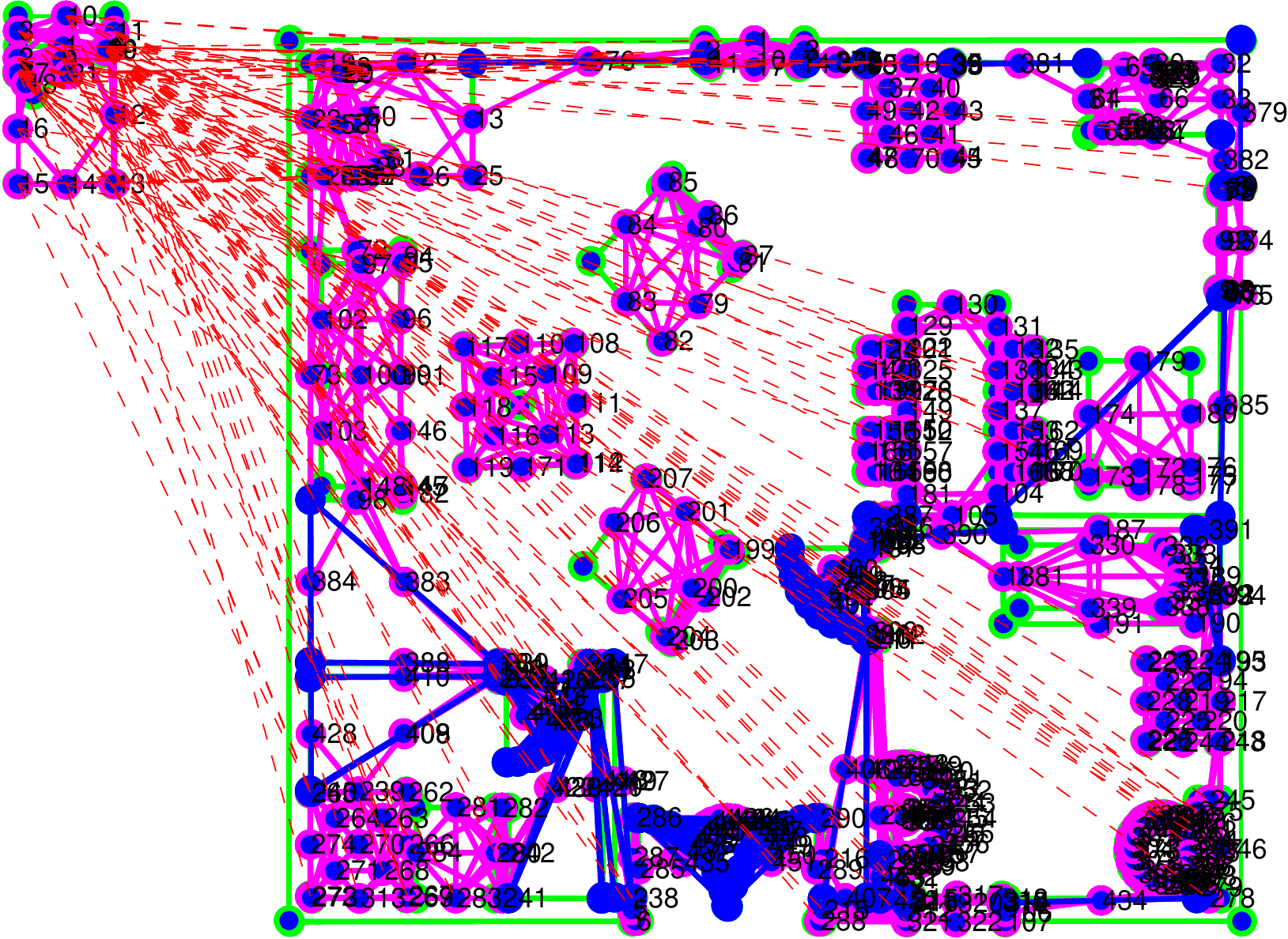}}
\hspace{0.5mm}
\subfloat[]{\includegraphics[width=0.32\textwidth,height=0.24\textwidth]{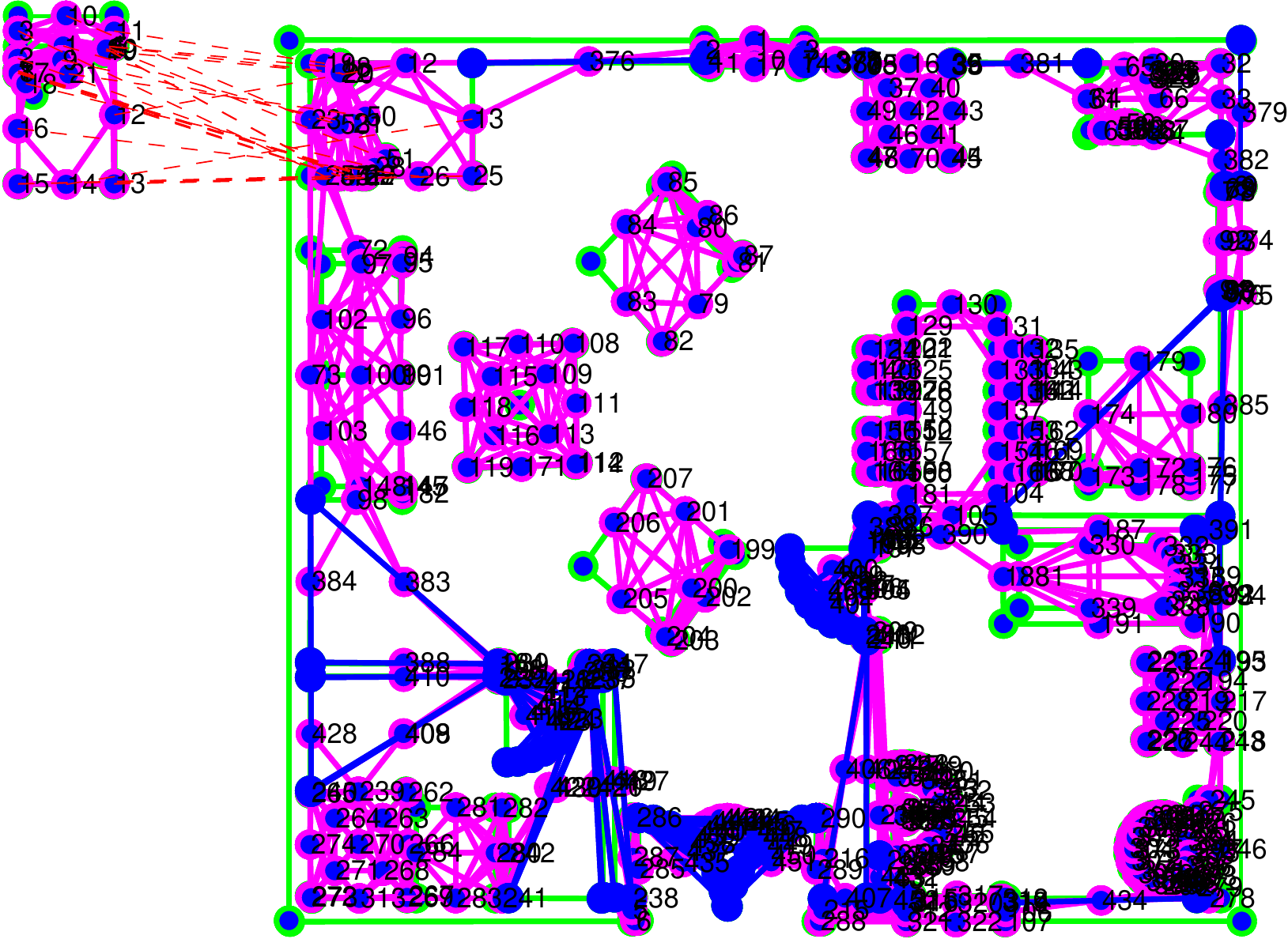}}
\hspace{0.5mm}
\subfloat[]{\includegraphics[width=0.32\textwidth,height=0.24\textwidth]{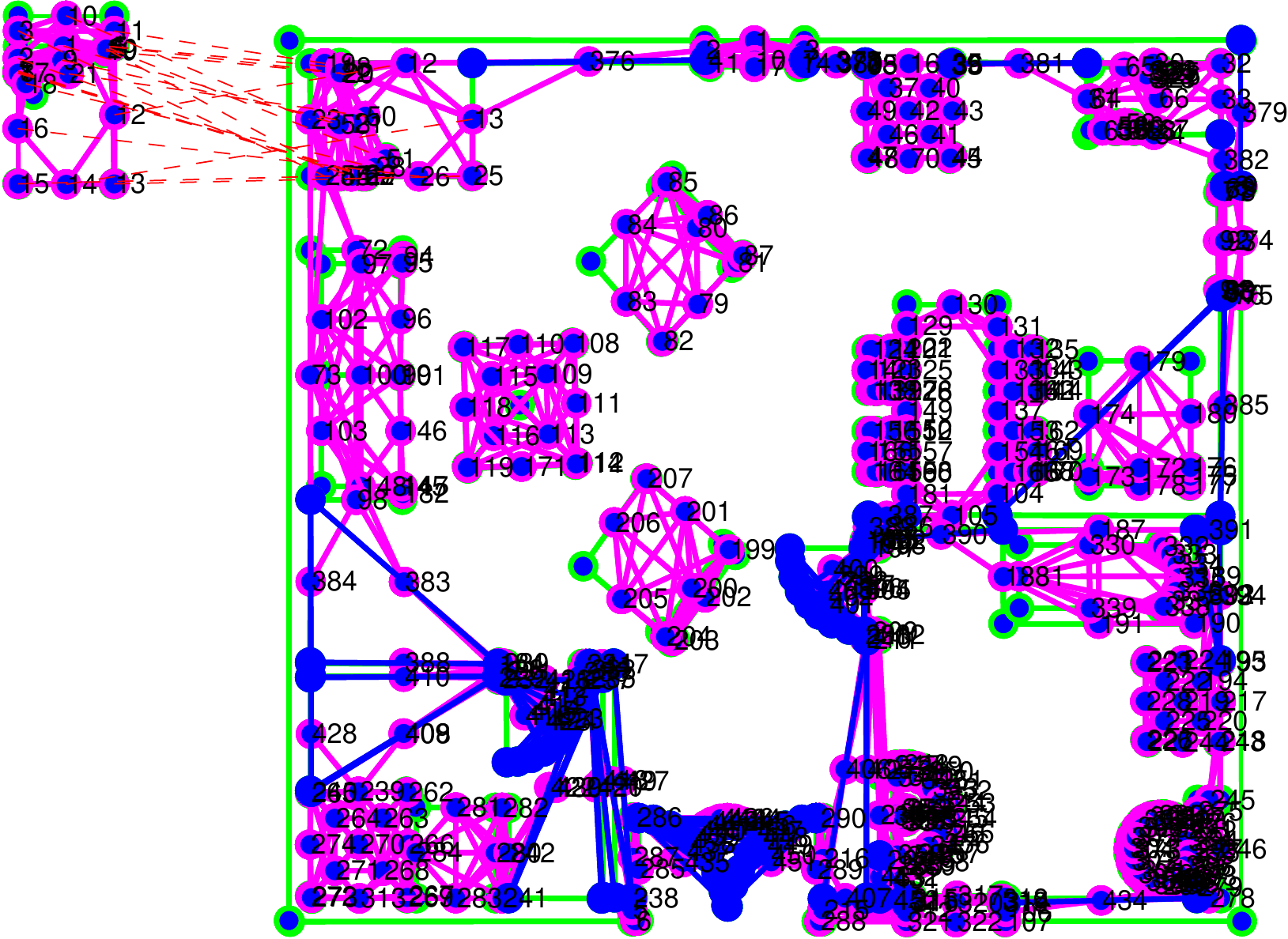}}
\caption{Robustness of contextual similarities in presence of outliers. The correspondences are shown based on the obtained linear programming responses \ie~without applying the discretization step: (a) pairwise similarities (PG-N), (b) random walk based contextual similarities (PG-R), (c) backtrackless walk based contextual similarities (PG-B). (Best viewed in pdf.)}
\label{fig:matching-bed}
\end{center}
\end{figure*}

In \fig{fig:matching-bed}, we have shown an example of node-node matching for three different scenarios. In that figure, the left sub figure shows the correspondences obtained by PG-N, the one in the middle shows the correspondences obtained by PG-R and the one in the right shows the same with PG-B. Here it is clear that the contextual similarities really enhance the truly matched nodes, as in the case of random, backtrackless walks the most of the obtained correspondences focus to the true matchings. After obtaining the correspondences, we performed a simple clustering to group the adjacent dual nodes to obtain a bounding box for comparing with the ground truth. A single bounding box is supposed to contain a single instance of a pattern graph.

%\subsubsection{Performance evaluation and results}
The decision whether a retrieved subgraph is true or false was done by overlapping the bounding boxes. This signifies: if $B$ be the rectangular bounding box for a retrieval and $T$ be the bounding box of its corresponding ground truth, $B$ is considered true if $\frac{B\cap T}{B\cup T}\geq 0.5$. For quantitative evaluation, we computed the usual metrics for evaluating a retrieval system, such as precision (\textbf{P}), recall (\textbf{R}), F-measure (\textbf{F}) and average precision (\textbf{AP}). For detailed definitions of all such metrics, we refer to the paper~\cite{Rusinol2009}. To get an idea about the time complexity we have also shown the meantime (\textbf{T}) of matching a pattern graph in a target graph. To have a comparison among the state-of-the-art symbol spotting methods, we considered five different methods. They are: (1) Symbol spotting with graph representation (SSGR)~\cite{Qureshi2007}, (2) Integer linear programming (ILPIso)~\cite{LeBodic2012}, (3) Fuzzy graph embedding (FGE)~\cite{Luqman2013}, (4) Serialized graphs (SG)~\cite{Dutta2013}, (5) Near convex region adjacency graph (NCRAG)~\cite{Dutta2013b}.

\begin{figure*}[!t]
\begin{center}
\subfloat[]{\label{sfig:pr-plot}\includegraphics[width=0.24\textwidth,height=3.4cm]{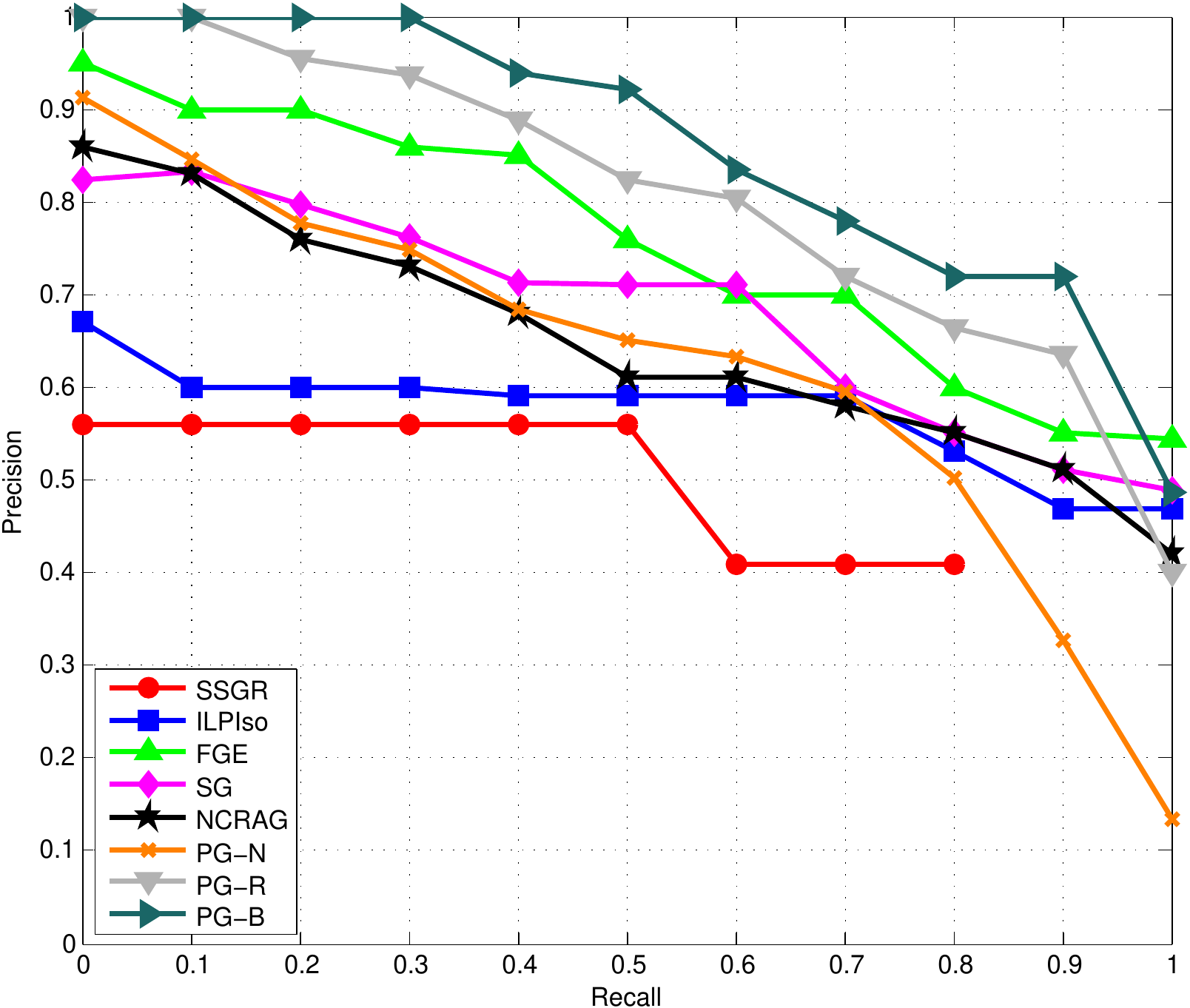}}
\hspace{0.5mm}
\subfloat[]{\label{sfig:roc-sink3}\includegraphics[width=0.24\textwidth,height=3.5cm]{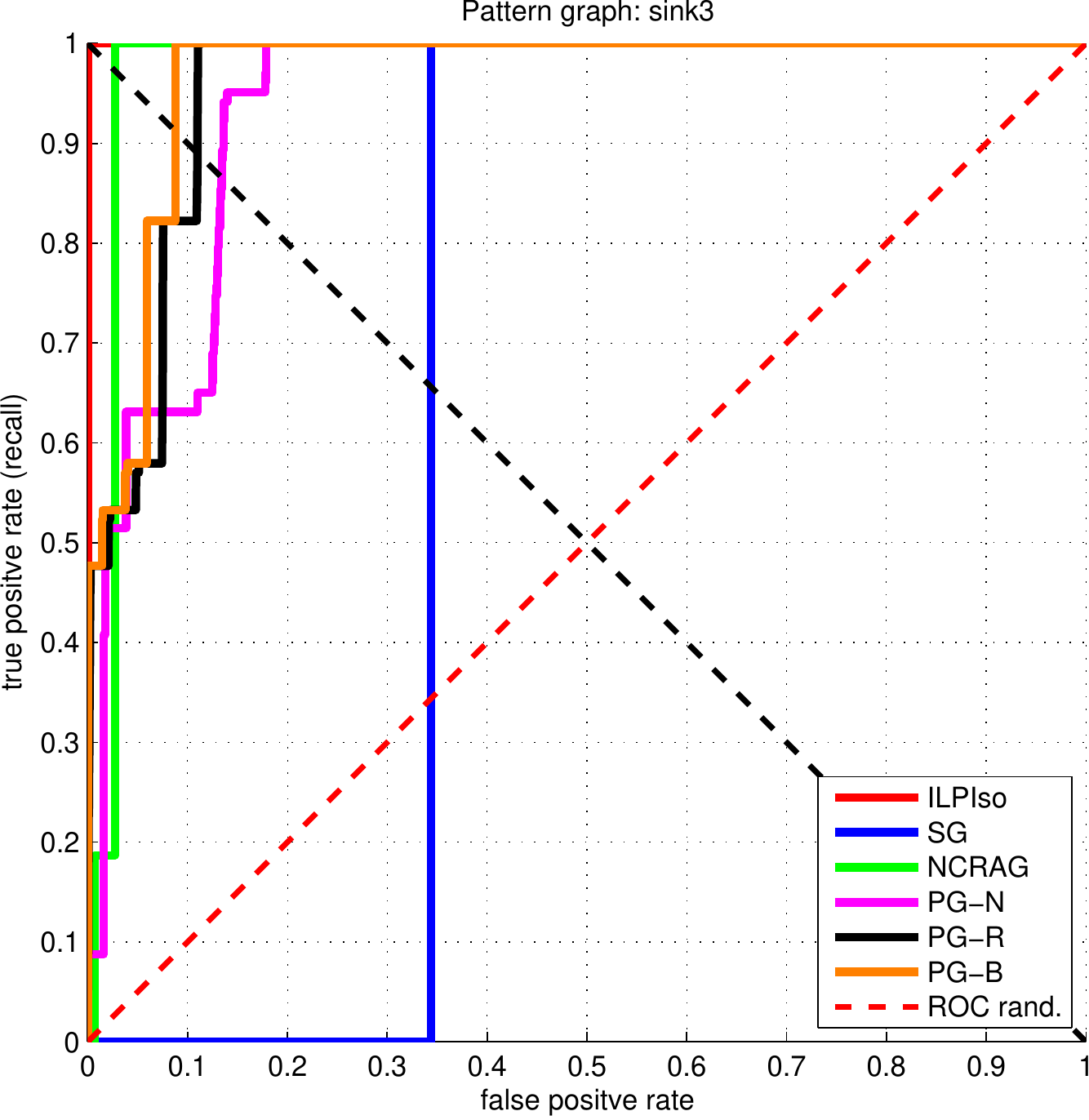}}
\hspace{0.5mm}
\subfloat[]{\label{sfig:roc-door1}\includegraphics[width=0.24\textwidth,height=3.5cm]{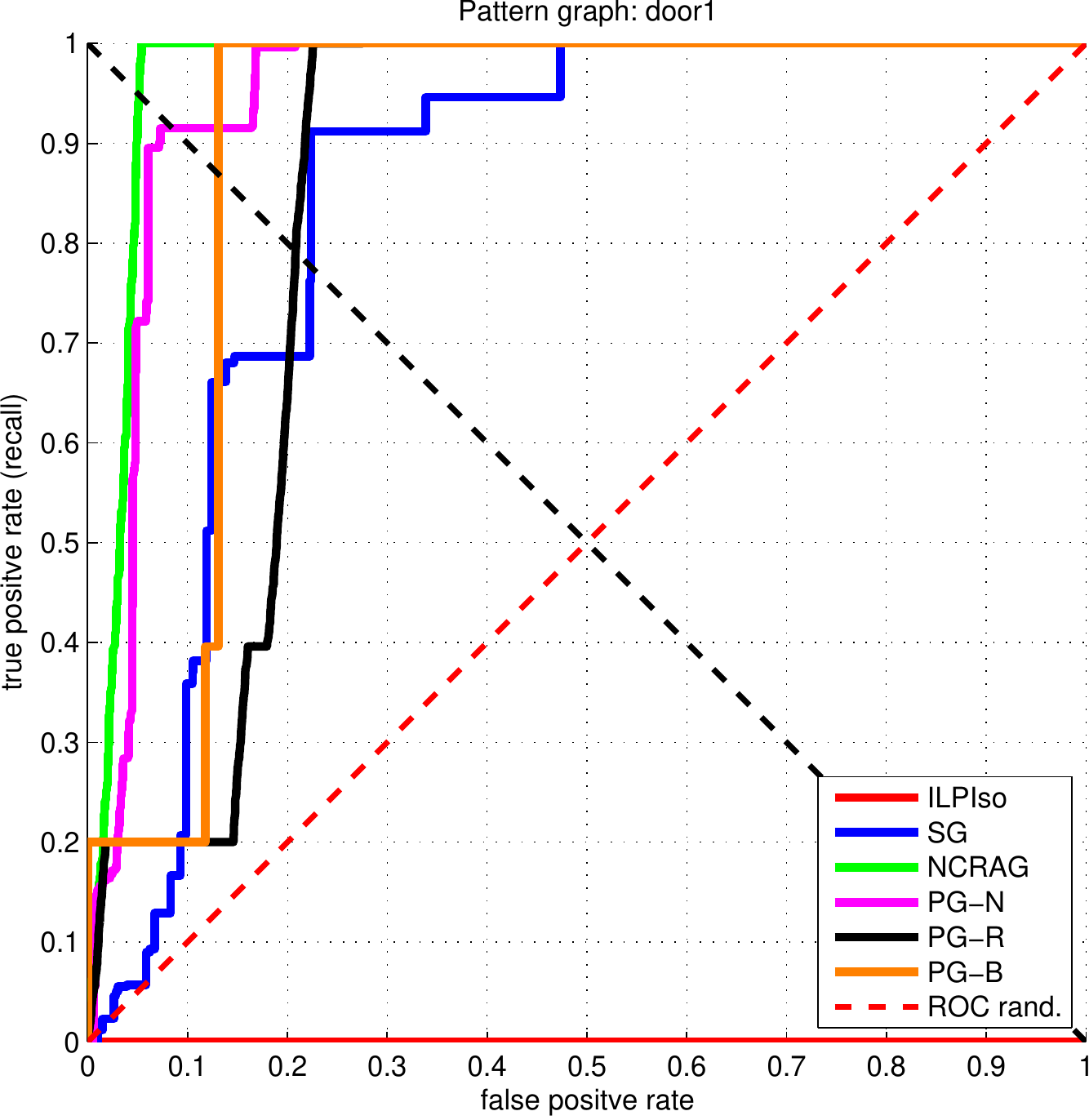}}
\hspace{0.5mm}
\subfloat[]{\label{sfig:roc-sofa1}\includegraphics[width=0.23\textwidth,height=3.5cm]{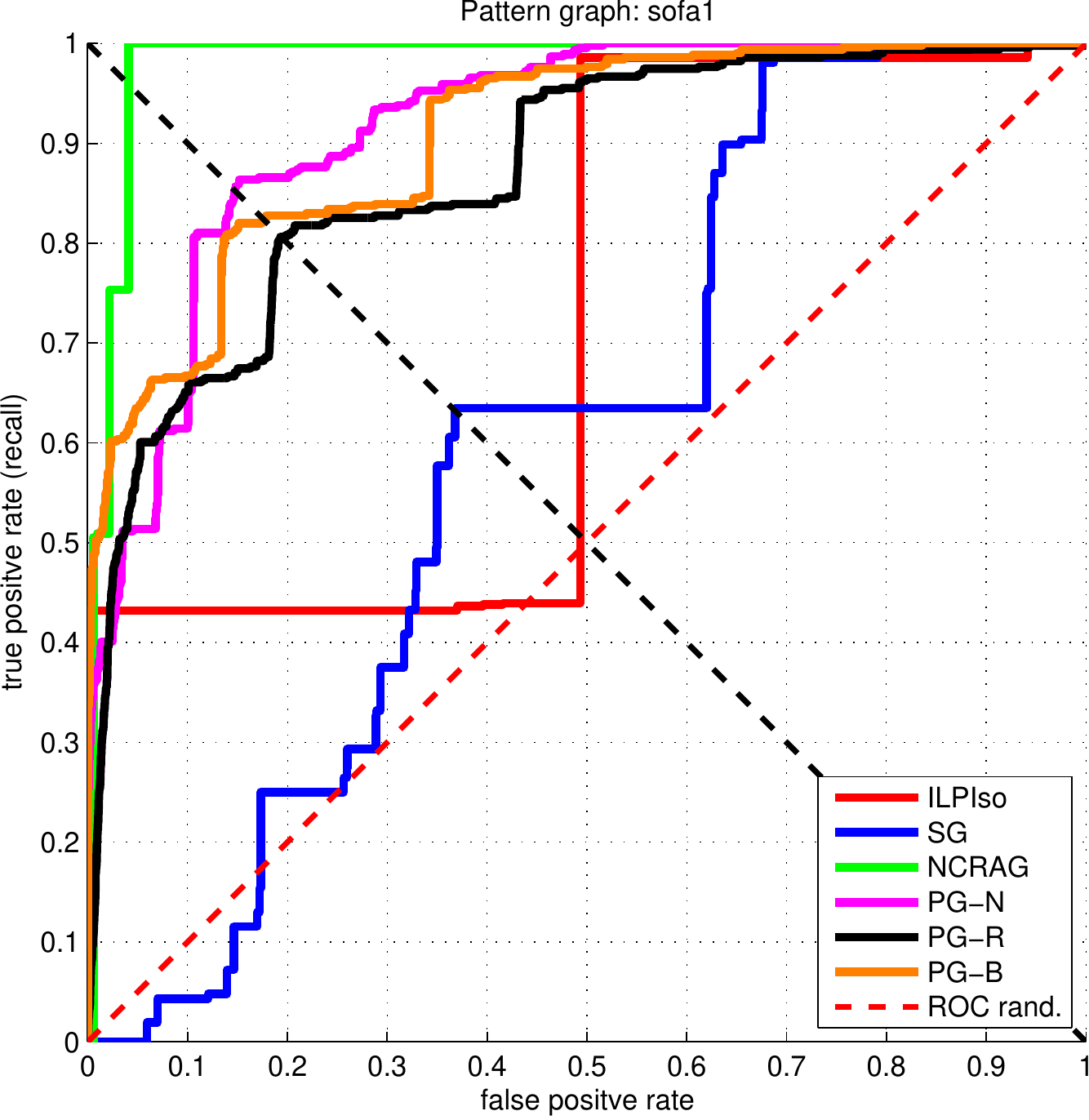}}
\caption{Performance of different methods on SESYD dataset: (a) overall precision-recall curves, (b) roc curves for \emph{sink3}, (c) roc curves for \emph{door1}, (d) roc curves for \emph{sofa1}. (Best viewed in pdf.)}
\end{center}
\end{figure*}

\begin{figure*}[!t]
\begin{center}
\subfloat[]{\includegraphics[width=0.32\textwidth]{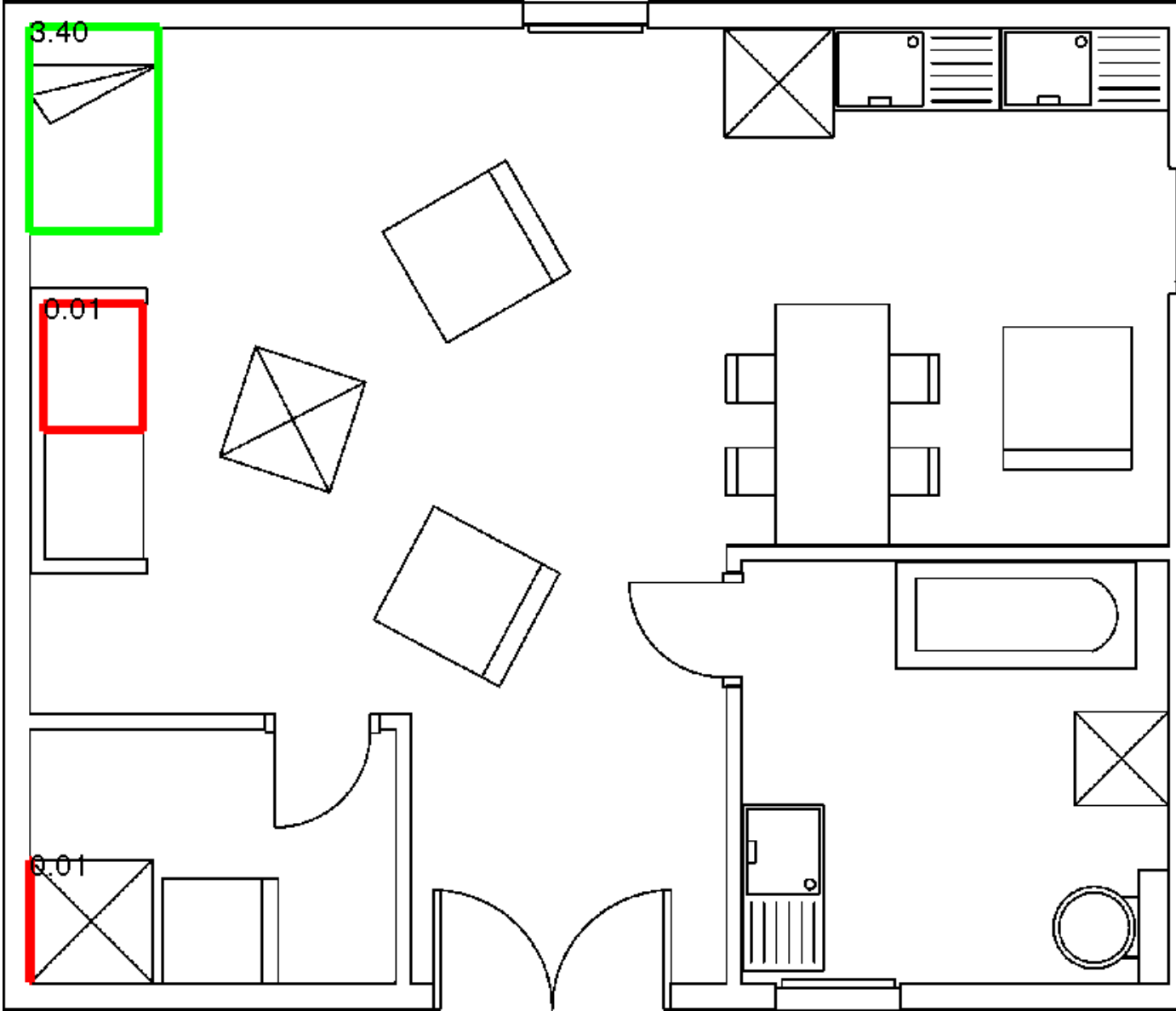}}
\hspace{0.5mm}
\subfloat[]{\includegraphics[width=0.32\textwidth]{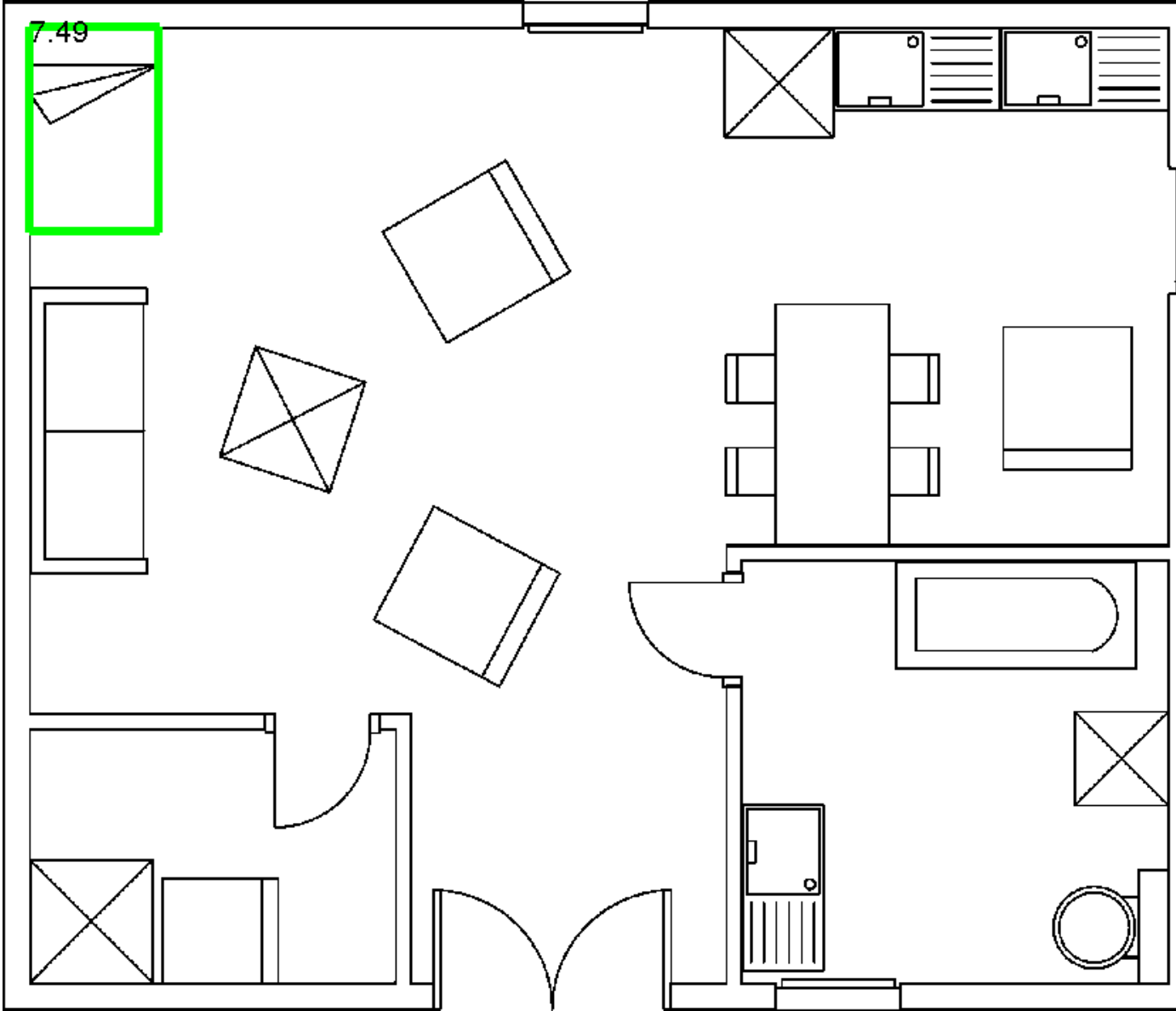}}
\hspace{0.5mm}
\subfloat[]{\includegraphics[width=0.32\textwidth]{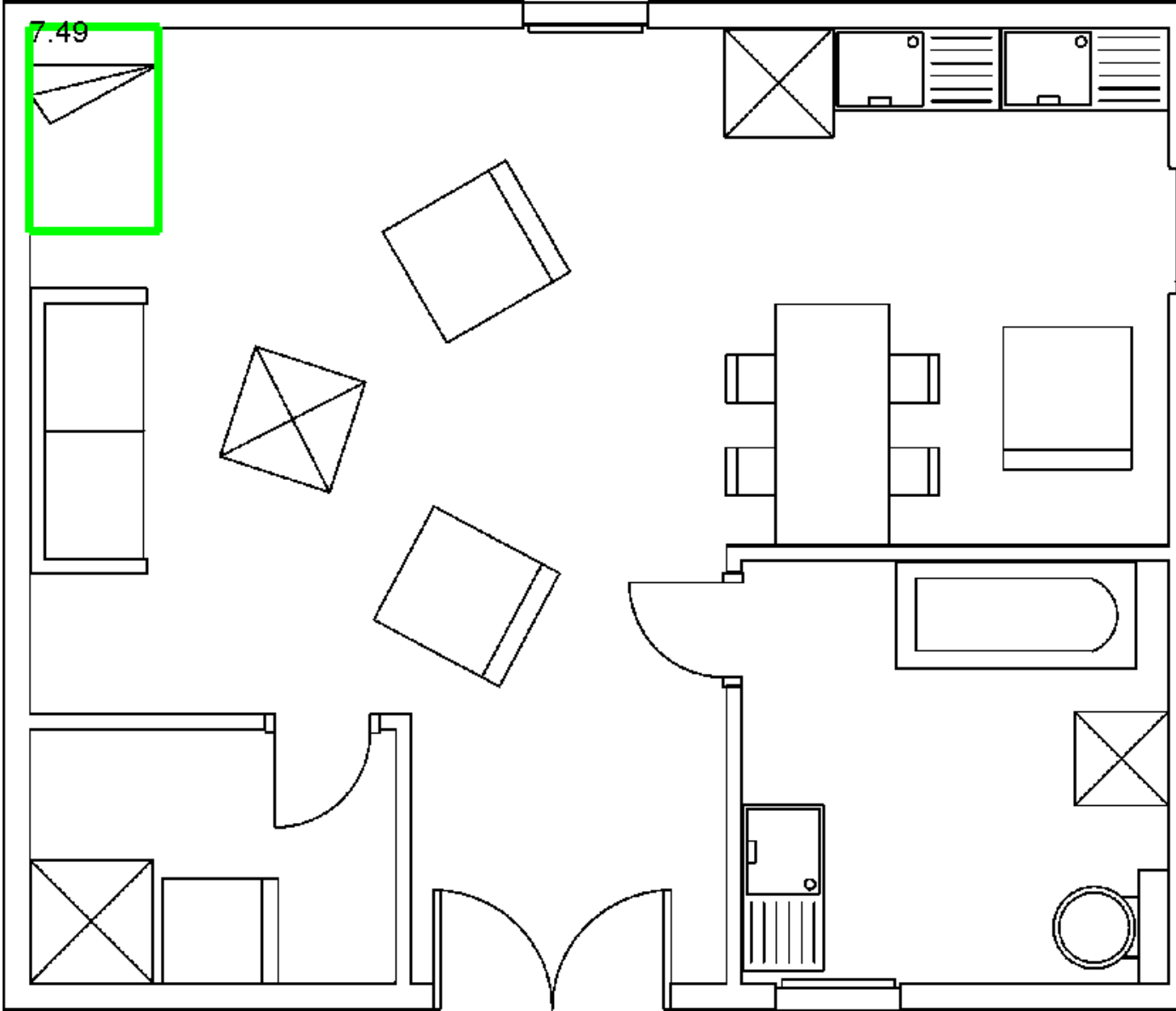}}
\caption{Symbol spotting: \emph{bed}: (a) PG-N, (b) PG-R, (c) PG-B. Green boxes are the true positives, the red ones are false positives. (Best viewed in pdf.)}
\label{fig:res-bed}
\end{center}
\end{figure*}

The quantitative results obtained by different methods are listed in \tab{table:res-method} and the precision-recall curves are shown in \fig{sfig:pr-plot}. From the results it is clear that the methods working with contextual similaritis (PG-R and PG-B) outperformed all the other methods. Moreover, there is a difference observed in the performance of PG-R and PG-B, and this phenomenon can be explained as a benefit of PG-B in the case of undirected graphs which eventually eliminates the totterings in the graph edges. Like most of the methods, our methods showed comparatively low performance for \emph{sink3} (\fig{sfig:roc-sink3}) and \emph{door1} (\fig{sfig:roc-door1}). This is because of the existence of \emph{sink2} and \emph{door2}, which respectively contain the previously mentioned pattern graphs. Also worse performance was obtained for \emph{sofa1} (\fig{sfig:roc-sofa1}) and this happened as this pattern frequently occurred in floorplans.

\begin{table}[h!]
\small
\centering
\caption{Comparative results with different existing methods}
\begin{tabular}{cccccc}
\toprule
\hline
\textbf{Methods} & \textbf{P} & \textbf{R} & \textbf{F} & \textbf{AveP} & \textbf{T}\\
\hline
SSGR~\cite{Qureshi2007}& 41.00 & 80.00 & 54.21 & 64.45 & -\\
ILPIso~\cite{LeBodic2012} & 65.44 & 58.23 & 59.11 & 57.75 & 27.29s\\
FGE~\cite{Luqman2013}& 56.00 & \textbf{100.00} & 71.79 & 75.50 & -\\
SG~\cite{Dutta2013}& 54.19 & 83.98 & 65.87 & 65.43 & \textbf{0.07s}\\
NCRAG~\cite{Dutta2013b} & 61.89 & 82.87 & 70.85 & 70.65 & 0.72s\\
PG-N & 61.60 & 72.81 & 64.94 & 60.21 & 53.60\\
PG-R & 69.90 & 84.95 & 78.78 & 81.10 & 33.43\\
PG-B & \textbf{70.56} & 86.29 & \textbf{80.10} & \textbf{82.98} & 33.37s\\
\hline
\end{tabular}
\label{table:res-method}
\end{table}

A qualitative result\footnote{Rest of the qualitative results are available on~\tt{\url{https://sites.google.com/site/2adutta/research}}} is shown in \fig{fig:res-bed}, where the subfigure on the left was obtained by PG-N, the middle one by PG-R and the right one by PG-B. In these figures, the bounding boxes with green border indicate the true positives and the ones with red border signify false positives. Each of the retrieved bounding boxes are also accompanied with a similarity value. The similarity value for each of the bounding boxes is the summation of all the similarities of matched nodes. These similarity values were used to rank the retrievals.
%------------------------------------------------------------------------------------------------------------------------------------------------------------------------------------------------------------
\section{Conclusions}
\label{sec:concl}
In this paper, we have proposed a random walk based approach on TPG to obtain contextual similarities between nodes of the operand graphs of TPG, that capture higher order information. In the next step, these contextual similarities are used to formulate subgraph matching procedure as a node and edge selection procedure in TPG, which is an optimization problem and has been solved with a linear programming relaxation. Since contextual similarities consider neighbour information of nodes/edges, this approach adds more discrimination power to nodes and edges. As a result, graph matching using linear programming turns out to be robust and efficient. We have presented a detailed experimental evaluation with varied difficulty and in most experiments our method performed better than the state-of-the-art methods. In general, it has been revealed that PG-B is more robust than PG-R, which is also supported by the proposal of Aziz~\etal~\cite{Aziz2013}. And both PG-B and PG-R perform better than PG-N, which proves the effectiveness of the proposal. Usually, low edge density of the operand graphs results in poor performance, which is well justified since the contextual similarities are generated by the propagation of pairwise similarities through the edges of TPG. So the absence of edges disturbs the similarity values.

Since the optimization problem is solved using linear programming, mostly in inexact cases (\eg~symbol spotting experiment) we do not get a one-one mapping between the nodes and edges. Instead we have performed a simple grouping to cluster the nodes and rank them depending on the total similarity. In future, motivated by Wang~\etal~\cite{Wang2013}, this approach can be extended by adding a density maximization based module that would discard the outliers and automatically cluster the inliers.



%% References


\bibliographystyle{elsarticle-num}
\bibliography{bibliography}






% \end{thebibliography}

\end{document}